\documentclass[10pt,twocolumn,letterpaper]{article}

\usepackage{3dv}

\usepackage{titling} % title in supp
\usepackage{chapterbib} % ref in supp
\usepackage[numbers]{natbib} % ref in supp
\usepackage{setspace} % spacing in supp

\usepackage{times}
\usepackage{epsfig}
\usepackage{graphicx}
\usepackage{amsmath}
\usepackage{amssymb}
\usepackage{stfloats}

\usepackage[pagebackref=true,breaklinks=true,colorlinks,bookmarks=false]{hyperref}

\newcommand{\NEW}[1]{#1}
\newcommand{\ignore}[1]{}

\newcommand{\pp}{{\bf p}}
\newcommand{\qq}{{\bf q}}
\newcommand{\rr}{{\bf r}}

\DeclareMathOperator*{\argmin}{arg\,min}

\threedvfinalcopy % *** Uncomment this line for the final submission

\def\threedvPaperID{56} % *** Enter the 3DV Paper ID here
\def\httilde{\mbox{\tt\raisebox{-.5ex}{\symbol{126}}}}

\ifthreedvfinal\fi

\begin{document}

\pagenumbering{gobble}
\title{Semi-Global Stereo Matching with Surface Orientation Priors}

\author{
Daniel Scharstein\\Middlebury College\\{\tt\small schar@middlebury.edu}
\and
Tatsunori Taniai\\RIKEN AIP\\{\tt\small tatsunori.taniai@riken.jp}
\and
Sudipta N.~Sinha\\Microsoft Research\\{\tt\small sudipta.sinha@microsoft.com}
}
\date{\vspace{-3ex}}
\maketitle
\thispagestyle{empty}

\begin{abstract}

Semi-Global Matching (SGM) is a widely-used efficient stereo
matching technique. It works well for textured scenes, but fails on
untextured slanted surfaces due to its fronto-parallel smoothness
assumption. To remedy this problem, we propose a simple extension,
termed SGM-P, to utilize precomputed surface orientation priors. Such
priors favor different surface slants in different 2D image regions or
3D scene regions and can be derived in various ways. In this paper we
evaluate plane orientation priors derived from stereo matching at a
coarser resolution and show that such priors can yield significant
performance gains for difficult weakly-textured scenes. We also
explore surface normal priors derived from Manhattan-world
assumptions, and we analyze the potential performance gains using
oracle priors derived from ground-truth data. SGM-P only adds a minor
computational overhead to SGM and is an attractive alternative to more
complex methods employing higher-order smoothness terms.

\end{abstract}

%%%%%%%%%%%%%%%%%%%%%%%%%%%%%%%%%%%%%%%%%%%%%%%%%%%%%%%%%%%%%%%%%%%%%%

\section{Introduction}

Semi-Global Matching (SGM) is a widely-used stereo matching technique
introduced by Hirschm\"{u}ller \cite{hirschmuller2005,hirschmuller2008}.
It combines the efficiency of local methods with the accuracy
of global methods by approximating a 2D MRF optimization problem with
several 1D scanline optimizations, which can be solved efficiently via
dynamic programming.  It has been shown that SGM is a special case of
message-passing algorithms such as belief propagation %(BP)
and TRW-T \cite{drory2014}.

SGM has had significant impact, and the method is widely used in
real-world applications, including 3D mapping,
robot and UAV navigation, and autonomous driving
\cite{hirschmuller2011,rabe-eccv2010, barry2015}.
SGM is also present in popular
computer vision libraries such as OpenCV and has been implemented in
hardware via FPGAs \cite{gehrig2009real} and on GPUs~\cite{banz2011}.

\begin{figure}[t]
\centering
\newcommand{\ww}{1.6in}
\begin{tabular}{@{}c@{~~}c@{}}
\includegraphics[width=\ww]{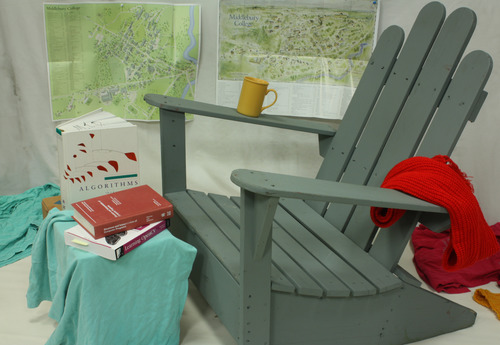}&
\includegraphics[width=\ww]{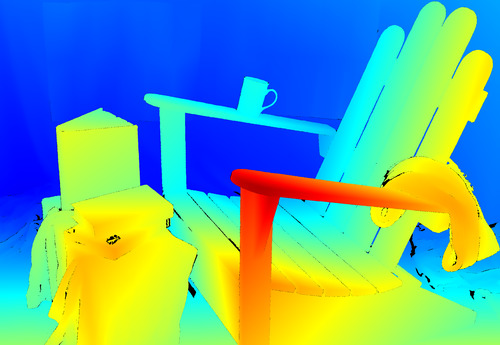} \\[-1mm]
\footnotesize (a) Input image &
\footnotesize (b) GT disparities \\[.5mm]
\includegraphics[width=\ww]{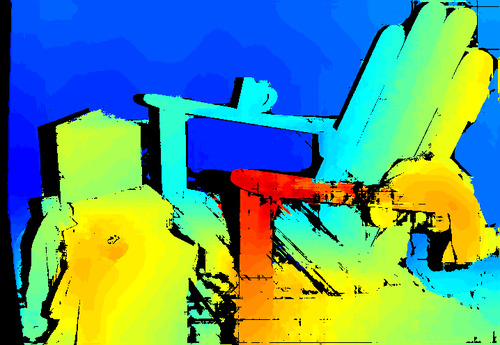}&
\includegraphics[width=\ww]{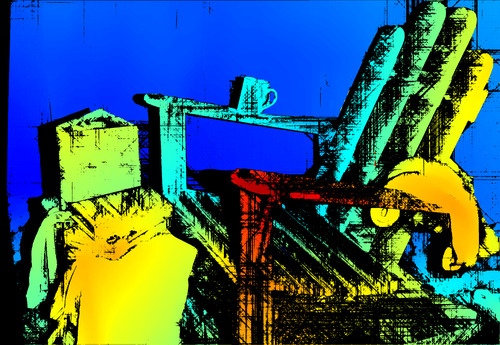} \\[-1mm]
\footnotesize (c) SGM, quarter resolution &
\footnotesize (d) SGM, full resolution \\[.5mm]
\includegraphics[width=\ww]{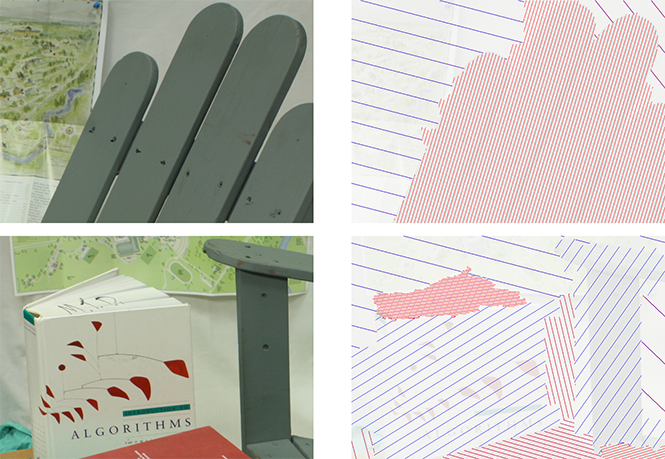}&
\includegraphics[width=\ww]{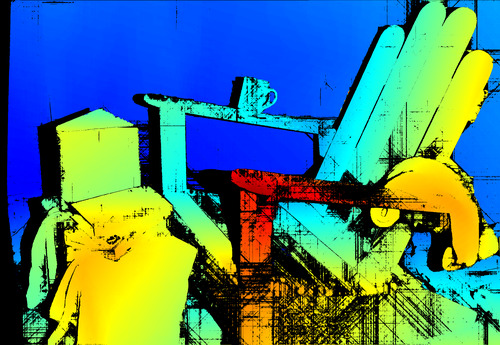} \\[-1mm]
%%\footnotesize (e) SGM-P, quarter resolution &
\footnotesize (e) Estimated orientation priors &
\footnotesize (f) SGM-P, full resolution\\[1mm]
\end{tabular}
\caption{
(a, b) Adirondack input image and ground-truth disparities.
(c, d) High-confidence standard SGM disparities at quarter and full resolution;
slanted surfaces cause problems at full resolution.
(e) Planar surface orientation priors derived from (c).
(f) Our SGM-P method uses these priors,
yielding high-confidence disparities
with significantly fewer holes on slanted surfaces.
\vspace{-3mm}
}
\label{fig:teaser}
\end{figure}

While the method works well for aerial imagery and textured outdoor scenes,
it works less well for indoor scenes with large
untextured regions.  The reason is that the algorithm employs a simple
first-order smoothness assumption that favors fronto-parallel
surfaces.  It therefore tends to hallucinate fronto-parallel patches
on weakly-textured slanted surfaces, which fail common
% left-right
consistency checks and result in large ``holes'' in the reconstructed
surface, in particular when matching high-resolution images
(Fig.~\ref{fig:teaser}d).

In this paper we propose a simple extension to the SGM algorithm,
SGM-P, that utilizes precomputed surface orientation priors.  Such
priors favor different surface slants in different regions of the
disparity space, and are implemented via local adjustments to SGM's
transition penalties.  The basic idea is to render the prior surface
in the 3D disparity space and store the locations of discrete disparity
steps in an \emph{offset image} (Fig.~\ref{fig:teaser}e).
SGM's smoothness term is then
modified so that the zero-cost surface follows these steps and thus
stays parallel to the prior surface.  Arbitrary surfaces, not just
planes, can be used as orientation priors, and our algorithm also
supports multiple surface priors at different depths via an
\emph{offset volume}.  Our method acts as a soft constraint during
matching and only adds a minor computational overhead.
In this paper we
demonstrate that even simple surface priors can yield significant
improvements in difficult indoor scenes containing slanted
surfaces with weak texture (see Fig.~\ref{fig:teaser}f).
Importantly, our experiments show that 
\NEW{in the absence of such
difficulties the performance never significantly decreases.}
%%our method does not negatively affect performance

The SGM-P algorithm is agnostic as to the source of the priors, which
could be computed in many different ways.  For instance, surface
priors can be derived from matched sparse feature points via
triangulation \cite{geiger2011efficient} or plane fitting
\cite{sinha-cvpr2014}.  Planes (or other parametric surfaces)
can also be fitted to disparities estimated at a lower resolution,
which is one method we explore in this paper.
Alternately, surface orientation priors could stem from domain
knowledge (e.g., expected ground plane orientation in autonomous
driving \cite{Konolige2008}),
from semantic analysis 
\cite{fouhey-iccv2013,fouhey-eccv2014,bansal-cvpr2016},
or from vanishing point analysis and Manhattan-world assumptions 
\cite{schindler2006,sinha-iccv2009,furukawa-cvpr2009,Lee2009,ramalingam2013},
which we also explore in this paper.
Finally,
surface priors could be derived from other sensors with
lower resolution (e.g., commodity depth cameras) to aid
high-resolution stereo matching in untextured regions.

%In the remainder of the paper we discuss related work
%(Section~\ref{sec:related}), present our method
%(Section~\ref{sec:algorithm}), evaluate its performance with various
%types of surface priors (Section~\ref{sec:experiments}), and conclude
%with ideas for future work (Section~\ref{sec:conclusion}).

\section{Related work}
\label{sec:related}

Stereo matching is one of the oldest and most-thoroughly studied
problems in computer vision \cite{Scharstein2002taxonomy,Brown03pami}.
% [other surveys?]
Methods can generally be categorized into local and global methods
\cite{Scharstein2002taxonomy}.  Both types of methods make smoothness
assumptions about the observed world; the former implicitely (e.g., by
aggregating a matching cost over a local window), and the latter
explicitly via a smoothness term that imposes a prior on the surfaces
in the world.  The simplest and most common smoothness assumption is
\emph{first order} and states that two neighboring pixels most likely
have the same depth.  This is assumed in both simple
window-based methods such as SSD and pixel-based global MRF
approaches \cite{boykov-pami2001}.
A first-order smoothness assumption introduces a fronto-parallel bias.
This is not a problem when there is sufficient texture in the scene, but
causes errors on untextured slanted surfaces, which is often
%particularly
problematic for indoor scenes (see Fig.~\ref{fig:teaser}d).

Many approaches have been proposed that can handle slanted surfaces.
Woodford et al.~\cite{Woodford-pami2009} show how second-order
smoothness terms can be efficiently optimized via QBPO.
% http://www.robots.ox.ac.uk/~ojw/2op/
% Global Stereo Reconstruction under Second Order Smoothness Priors
% O. J. Woodford, P. H. S. Torr, I. D. Reid, A. W. Fitzgibbon
% CVPR 2008 Paper (pdf) [best paper award],
%
% Second-order priors on the smoothness of 3D surfaces are a better
% model of typical scenes than first-order priors. However, stereo
% reconstruction using global inference algorithms, such as graph-cuts,
% has not been able to incorporate second-order priors because the
% triple cliques needed to express them yield intractable
% (non-submodular) optimization problems.
%
% This paper shows that inference with triple cliques can be effectively
% optimized. Our optimization strategy is a development of recent
% extensions to a-expansion, based on the QPBO algorithm. The strategy
% is to repeatedly merge proposal depth maps using a novel extension of
% QPBO. Proposal depth maps can come from any source, for example
% fronto-parallel planes as in a-expansion, or indeed any existing
% stereo algorithm, with arbitrary parameter settings.
Li and Zucker~\cite{li2010} derive smoothness models for slanted and
curved surfaces using differential geometry.
% Surface geometric constraints for stereo \cite{li-emmcvpr2005,li-cvpr2006}
% http://cs-www.cs.yale.edu/homes/li-gang/research/SurfaceStereo/index.html
% http://www.cs.yale.edu/homes/li-gang/publication/emmcvpr05.pdf
Bleyer et al.~propose surface stereo~\cite{bleyer-cvpr2010} and
object stereo~\cite{bleyer-cvpr2011} algorithms in which the scene is modeled
with planes or splines, and
% In a different line of work, Bleyer et al.~proposed
PatchMatch
stereo~\cite{bleyer-bmvc2011}, in which local estimates of disparities and
surface slant are propagated to neighboring regions.
%locations.
Sinha et al.~\cite{sinha-cvpr2014} run local
plane sweeps around
%surrounding
disparity planes estimated from sparse feature matches.

Plane-sweep stereo with a preferred plane orientation was proposed by Collins
\cite{Collins96}, and subsequently extended to multiple plane orientations,
Manhattan-world scenes, and piece-wise planar scenes
\cite{gallup-cvpr2007,furukawa-cvpr2009,sinha-iccv2009,Kowdle2012}.

A more recent trend is to formulate stereo matching using continuous
MRF frameworks~\cite{Yamaguchi2012}. While PatchMatch stereo
\cite{bleyer-bmvc2011} was a greedy algorithm, follow-up work such as
PMBP~\cite{Besse2014} incorporates regularization.  Other work employs
MRFs with continuous labels, using fusion moves for
optimization~\cite{Olsson2013,Taniai2014}.
\NEW{Several recent stereo algorithms
\cite{hadfield-iccv2015,hane-cvpr2015,zhang-icra2017} use surface normal priors
derived from single
images~\cite{fouhey-iccv2013,ladicky-eccv2014,fouhey-eccv2014,bansal-cvpr2016}
These methods utilize continuous optimization
and require minimization techniques such as primal-dual methods or
linear programming.  These ideas cannot be directly incorporated into
SGM or another discrete MAP inference framework.  Our proposed
algorithm offers a simple and efficient alternative to such complex
approaches, and contributes a practical way of imposing surface
orientation priors amenable to discrete optimization.}

In the context of SGM, several improvements have been proposed.  The
CSGM method by Hirschm\"{u}ller \cite{hirschmuller-cvpr2006} estimates
the disparities in untextured regions by fitting planes to adjacent
textured pixels.  In contrast to such post-processing, our method
incorporates orientation priors during the matching.
\NEW{Hermann et al.~\cite{hermann2009} suggest an approximate second-order
smoothness term for SGM but do not demonstrate a clear performance gain.}
%%%\Daniel{Not sure what to say here, maybe don't cite it at all?}
%
A hierarchical approach to SGM \cite{rothermel2012,wenzel2013} aims
to reduce ambiguities and runtime by restricting the search range
based on SGM results computed at a lower resolution.  We use a
similar idea as one possible mechanism to derive surface
orientation priors.  While hard constraints such as search-range
reduction can cause fine detail to be missed, in our case we only use
the result from a coarser resolution to obtain a soft constraint on surface
orientations.

%http://www.ifp.uni-stuttgart.de/publications/2012/Rothermel_etal_lc3d.pdf
%http://www.ifp.uni-stuttgart.de/publications/phowo13/080Wenzel.pdf

Finally, our priors could also be added to other recent modifications
of SGM, for instance MGM \cite{facciolo2015}, which integrates results
from multiple directions.  Similarly, our technique is orthogonal to
recent advances in matching cost learning by CNNs
\cite{Zbontar2016} as we show in experiments below.

%Facciolo et al.~\cite{facciolo2015} propose MGM, a
%modification of the SGM algorithm that operates not only along 1D
%lines, but also integrates results from the perpendicular direction.
%Our priors could also be added to their technique.

%http://www.bmva.org/bmvc/2015/papers/paper090/paper090.pdf

\section{Algorithm}
\label{sec:algorithm}

We first review the SGM algorithm, then describe our proposed extension
%which we call [DS: we already introduced the term ealier]
SGM-P.

%%%%%%%%%%%%%%%%%%%%%%%%%%%%%%%%%%%%%%%%%%%%%%%
\begin{figure*}[t]
\centering
\includegraphics[width=\linewidth]{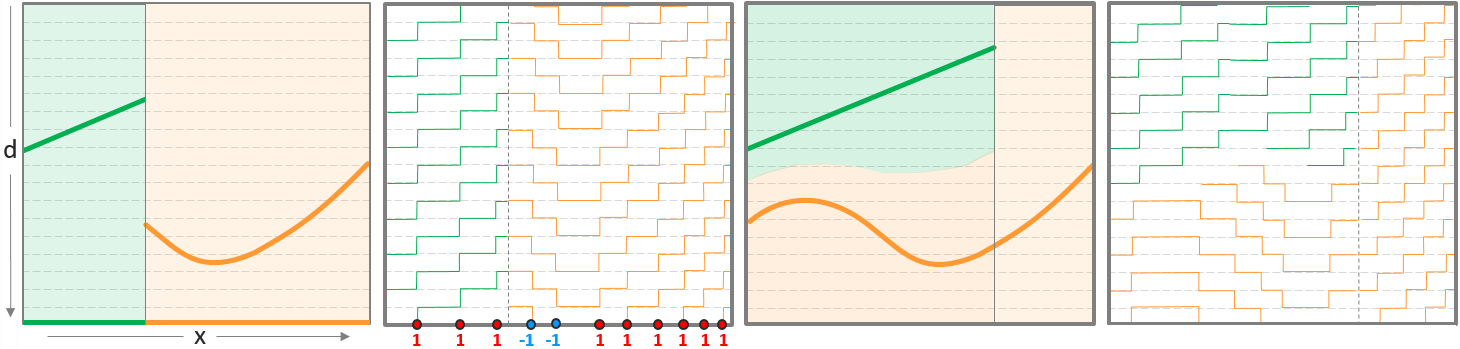}\\[-1mm]
\centerline{\small ~~ \hfill (a) \hfill \hfill (b) \hfill \hfill (c) \hfill \hfill (d) \hfill}
\caption{Illustration of SGM-P's smoothness term:  An $x$-$d$ slice of the
disparity volume with two prior surfaces (green and orange line)
whose orientations we want to encourage, and their
respective regions of influence (shaded).
(a) 2D orientation priors extend across all disparities. (b) Rasterized version;
the disparity jumps $\pm 1$ (red and blue circles) do not depend on $d$
and can be stored in an \emph{offset image}. (c) 3D priors allow
multiple surface hypotheses per pixel (here, two on the left and one on the right).
Each point is influenced by its closest surface in the $d$ direction, so
surfaces define Voronoi cells. (d) When multiple surfaces are present, disparity jumps
vary with $d$ and define an \emph{offset volume}.  The offsets are computed
for each surface segment within its respective Voronoi cell.}
\label{fig:mainidea}
\end{figure*}

\subsection{SGM}

The Semi-Global Matching (SGM) algorithm \cite{hirschmuller2008}
is an efficient technique for approximate energy minimization of a 2D
Markov Random Field (MRF),
\begin{equation}
E(D) = \sum_{\pp} C_\pp(d_\pp)
+  \sum_{\pp,\qq \in \cal{N}} V(d_\pp, d_\qq) ,
\label{eqn:mrf}
\end{equation}
where $C_\pp(d)$ is a unary data term that represents the cost
of matching pixel $\pp$ at disparity
$d \in \cal{D} =$$ \{d_{\min}, \ldots, d_{\max}\}$,
and $V(d, d')$ is a pairwise smoothness term that
penalizes disparity differences between neighboring pixels.
Specifically, $V$ implements a first-order smoothness assumption,
\begin{equation}
V(d, d') = \left\{
\begin{array}{ll}
0   & \mbox{if } d = d' \\
P_1 & \mbox{if } |d - d'| = 1 \\
P_2 & \mbox{if } |d - d'| \geq 2 .
\end{array}
\right.
\label{eqn:smoothness}
\end{equation}
Instead of minimizing the 2D MRF,  %%%(\ref{eqn:mrf}),
which is NP-hard,
SGM efficiently minimizes a 1D version of Eqn.~\ref{eqn:mrf} along 8
cardinal directions $\rr$ via dynamic programming~\cite{hirschmuller2008}.
For each direction $\rr$, SGM computes an aggregated matching cost
$L_\rr(\pp, d)$ recursively defined from the image boundary:
\begin{equation}
L_\rr(\pp, d) = C_\pp(d) + \min_{d' \in \cal{D}}(L_\rr(\pp -\rr, d')+ V(d, d')).
\end{equation}
The 8 aggregated costs are summed at each pixel, yielding an aggregated cost volume
\begin{equation}
S(\pp,d) = \sum_\rr L_\rr(\pp,d)
\end{equation}
whose per-pixel minima are chosen as the winning disparities
\begin{equation}
d_\pp = \argmin_d S(\pp, d).
\end{equation}

Drory et al.~\cite{drory2014} observe that the sum of the 8 individual
minima of $L_\rr(\pp,d)$ is a lower bound on the minimum of the
aggregated cost $S(\pp, d)$ at each pixel $\pp$, and
%%that an uncertainty measure $U_\pp$ can be defined
define an uncertainty measure $U_\pp$
as the difference between the two:
\begin{equation}
U_\pp =  \min_d \sum_\rr L_\rr(\pp,d) -  \sum_\rr \min_d L_\rr(\pp,d).
\label{eqn:uncertainty}
\end{equation}

The intuition is that $U_\pp$ will be zero at locations where the 8
minimum-cost paths agree, e.g., in textured regions where incorrect
disparities have high unary costs $C_\pp$.  In untextured regions,
however, multiple disparities will have similar unary costs, and the 8
individual minima of $L_\rr$ will likely occur at different
disparities, in particular on slanted surfaces.  We use $U_\pp$ in our
experiments to plot disparity error as a function of uncertainty,
and also in Fig.~\ref{fig:teaser} to select high-confidence matches.

\subsection{SGM-P}

In order to utilize surface priors,	
the basic idea is to modify SGM's smoothness penalties to favor
surfaces with the expected surface slant.  The problem is that we
cannot represent fractional surface slants in algorithms that use
discrete disparities, such as SGM.  Thus, the key idea is to
\emph{rasterize} the disparity surface, i.e., render it in the 3D pixel grid,
and record the locations of the steps in the discretized disparity
values.  At these locations we then shift SGM's smoothness penalties
$V$ so that the zero-cost transitions coincide with these steps.
See Fig.~\ref{fig:mainidea} for illustration.
We first discuss the case where we have only one orientation
prior per pixel.

%%\subsection{2D disparity surface orientation prior}
\subsection{2D orientation priors}  % DS - shorter is better
\label{sec:2Dprior}

Assume we are given a real-valued disparity surface prior $S$ whose
orientation at any given pixel $\pp$ we would like to encourage across
all possible disparities (Fig.~\ref{fig:mainidea}a).

Let $\rr$ be the current ``sweeping direction'' of SGM.  Given a pixel
$\pp$ and its successor $\pp' = \pp + \rr$, we rasterize the surface
$S$ to integer disparities
\begin{equation}
\hat{S}(\pp) = \mbox{round}(S(\pp))
\label{eq:round}
\end{equation}
and compute the discrete disparity steps (or \emph{jumps})
\begin{equation}
j_\pp = \hat{S}(\pp') - \hat{S}(\pp).
\label{eq:jump_p}
\end{equation}
We replace the original smoothness penalty $V$ with a
new function $V_S$ that incorporates the disparity jumps:
\begin{equation}
V_S(d_\pp, d_\pp') = V(d_\pp + j_\pp, d_\pp').
\end{equation}
At pixels where the value of $j$ is nonzero, $V_S$ favors taking that
disparity step and encourages the disparity surface to stay parallel to $S$.
We can efficiently compute $V_S$ by storing the jumps $j$ in
an \emph{offset image} (Fig.~\ref{fig:mainidea}b).
Since the jumps depend on the direction $\rr$, four different offset
images are needed, one for each pair of opposing directions.

However, we do not need to precompute and store all four offset images simultaneously.
Instead, we only need to store $\hat{S}$. Before running
scanline optimization on each pair of opposing directions, we generate the
associated offset image using Eqns.~\ref{eq:round} and~\ref{eq:jump_p}.
When reversing the direction, we flip the signs of the offsets.

\subsection{3D orientation priors}
\label{sec:3Dprior}

We now relax the requirement of a single orientation prior per pixel
and allow multiple overlapping surface hypotheses at different depths.
As before, each surface should only act as an \emph{orientation} prior,
but should influence all nearby points (Fig.~\ref{fig:mainidea}c).
Assume that at pixel $\pp$ we have $K$ disparity
surfaces $\{S^{\pp}_k\}, k = 1 \ldots K$. For a given disparity $d$,
we find the closest surface in terms of disparity
\begin{equation}
\tilde{S}^{\pp}_d = \argmin_k |S^{\pp}_k(\pp) - d \big|.
\end{equation}
Then, we rasterize it to integer disparities
\begin{equation}
\hat{S}_d(\pp) = \mbox{round}(\tilde{S}^{\pp}_d(\pp)),
\end{equation}
and again compute the discrete disparity jumps
between adjacent pixels. This time, however, they depend on $d$:
\begin{equation}
j_\pp(d) = \hat{S}_d(\pp') - \hat{S}_d(\pp).
\end{equation}
The new smoothness penalty term now becomes
\begin{equation}
V_S(d_\pp, d_\pp') = V(d_\pp + j_\pp(d_\pp), d_\pp').
\end{equation}
Fig.~\ref{fig:mainidea}d illustrates this disparity-dependent orientation
prior.
In order to compute $V_S$ efficiently, we store the precomputed
values of $\tilde{S}^{\pp}_d$ for all $\pp$ and $d$ in an auxiliary
volume. This can be seen as a 1D discrete Voronoi diagram for each
pixel along the disparity axis in the volume. The values in each
column can be computed efficiently with a forward scan followed by a
backward scan after rendering all disparity surfaces into the column
at this pixel.

\begin{figure}[t]
\centering
\includegraphics[width=0.48\linewidth]{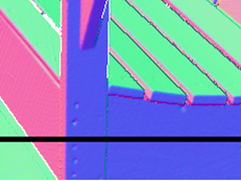}
\hspace{1mm}
\includegraphics[width=0.48\linewidth]{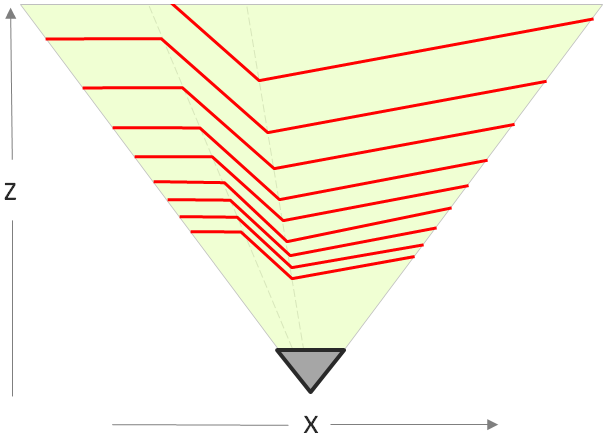}\\
\small (a) \hspace{3.5cm} (b)\\
\includegraphics[width=0.48\linewidth]{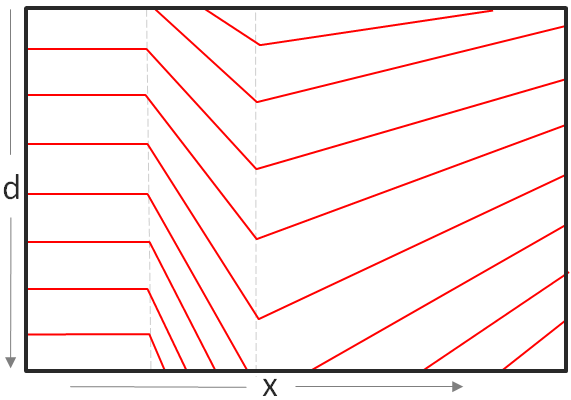}
\hspace{1mm}
\includegraphics[width=0.48\linewidth]{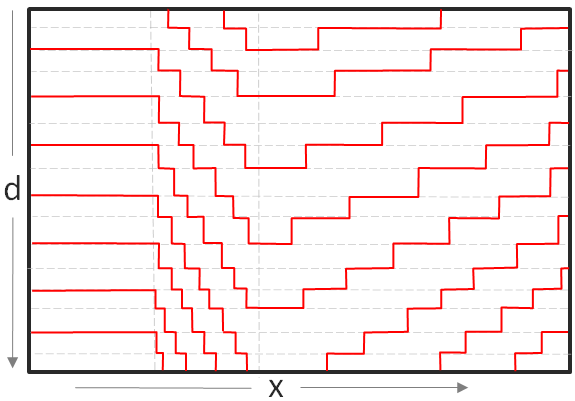}\\
\small (c) \hspace{3.5cm} (d)\\
\caption{Converting surface normals to disparity orientation priors.
(a) Color rendering of a surface normal map with a
scanline spanning three piecewise-planar surfaces.
(b) Integrating normals under perspective projection gives rise to
parallel surfaces with varying scale and depth.
(c) Converting to disparity space, viewing rays become parallel, and surface
slant now varies with disparity.
(d) The offset volume encodes all possible surface orientations;
note that disparity steps are no longer aligned vertically.
%\Daniel{I noticed we use (u, v) in the text, but still use x
%in (c), (d), and also in Fig 2. But I don't really want to change the x
%in the figures... maybe no one will notice.}
}
\label{fig:surfacenormals}
\end{figure}

\subsection{Surface normal priors}
\label{sec:normals}

So far we have considered using disparity surfaces as orientation
priors.  We now focus on the case where we are given a surface normal
map as prior, obtained for instance via photometric stereo
\cite{du2011} or from
%%semantic analysis \cite{bansal-cvpr2016}.
Manhattan-world priors \cite{Lee2009}.
Such
normal maps can be used in our SGM-P algorithm, though the situation
is more complex than one might expect.
First, we cannot use orientations directly, since \mbox{SGM-P}
requires an actual surface that can be rasterized.  Thus, the normal
map
must be integrated.  Second, we need to distinguish between scene
space (in Euclidean world coordinates) and disparity space.  While a
surface normal map can be considered a 2D prior since it encodes a
single surface orientation per pixel, this orientation is given in
scene space.  As we show below, when converting to disparity space
under a perspective projection model, the surface orientation becomes
depth-dependent and results in a 3D disparity orientation prior,
requiring an offset volume representation.  See
Figure~\ref{fig:surfacenormals} for illustration.

%%We begin by deriving the precise mathematical relationship between surface
We now derive the relationship between surface
normals in scene coordinates and the orientation of disparity surfaces.
%
%\vspace{2mm}
%\noindent \textbf{Surface normal-based disparity planes.}
%
Given the surface normal vector $(n^p_x, n^p_y, n^p_z)^T$ at pixel
$\pp$, the equation of the tangent plane of the surface at $\pp$
in scene space coordinates $(x,y,z)$
%%^T \in \mathbb{R}^3$,
is
%\begin{equation}
$
n^p_x x + n^p_y y + n^p_z z = h^p.
$
%\label{eq:tangentplane}
%\end{equation}
Here, $h^p$ encodes the plane's unknown depth.
Under perspective projection we have $x = uz/f$ and $y = vz/f$
for image coordinates $(u,v)$ and a camera
at the origin
with focal
length $f$.
% located at the origin with identity rotation.
Substituting these values
%into Eqn.~\ref{eq:tangentplane}
%%and solving for $z$
we obtain
\begin{equation}
z = h^p f / \big(n^p_x u + n^p_y v + f n^p_z \big).
\label{eq:z}
\end{equation}
For stereo pairs we have $z = bf/d$, where $b$ and $d$ are the baseline and
disparity respectively. Substituting $z$ into Eqn.~\ref{eq:z} we
obtain the disparity plane equation
\begin{equation}
d(u,v) = \frac{b}{h^p}\big(n^p_x u + n^p_y v + f n^p_z \big).
\label{eq:dplane}
\end{equation}
Note that the disparity plane orientation depends on $h^p$,
which encodes the depth of the tangent plane in scene space.
Therefore, a scene plane with fixed orientation but
unknown depth yields a family of disparity planes whose orientation
depends on the associated disparity.
%% (Fig.~\ref{fig:surfacenormals}c).

%\vspace{2mm}
%\noindent \textbf{Obtaining Disparity surfaces from normal maps.}

%\Daniel{Shorten the following discussion and integrate
%Manhattan approach here?}

%As mentioned,
In order to use surface normal priors in SGM-P, they
first need to be integrated into a surface.  We do this integration in
scene space under a perspective projection model using a least-squares
approach~\cite{tankus2005}
(see the supplementary materials for more details).
The result is a $z$-surface in scene space
coordinates, initially at an arbitrary depth.
%
%\Daniel{I got rid of the details (grid cells/ discontinuities).
%Is the description still accurate for Manhattan?}
%
%We implement the 2D integration step using a sparse solver employing
%conjugate gradients.
%We divide the image into a coarse grid and
%independently integrate a surface in each grid cell, arbitrarily
%fixing one depth value in each cell.
%
%In order for the integration to succeed, it is crucial that the
%discontinuities in the normal map are known. Otherwise any integration
%method, including our least-squares approach, will not produce a
%locally accurate $z$-surface.
%
%Given the integrated $z$-surface, we now
%convert it into a family of $d$-surfaces in the disparity volume using
%the perspective model.  We do this by scaling the original integrated
%$z$-surface
We scale this surface by an appropriate sequence of
% increasing
scale factors (Fig.~\ref{fig:surfacenormals}b)
and convert to $d$ to arrive at roughly
equally-spaced $d$-surfaces covering the full disparity range
(Fig.~\ref{fig:surfacenormals}c).
Finally, we construct an offset volume from this family of disparity
surfaces as described in the previous section, resulting in
a 3D orientation prior with varying disparity surface slants.
% for SGM-P
(Fig.~\ref{fig:surfacenormals}d).

\section{Experiments}
\label{sec:experiments}

We now demonstrate the utility of our new algorithm by comparing a
baseline SGM implementation with various version of SGM-P employing
different types of priors.  For a fair comparison we use the same
matching cost and smoothness weights across all versions of the
algorithms.

To compensate for global and local rectification errors, we first
robustly fit a global model $y' = ay+b$ to matched feature points and
warp the right image accordingly before computing the matching costs.
During matching, for each horizontal disparity, 
we evaluate matching costs corresponding to vertical disparities of \{-1, 0, +1\}
pixels and select the smallest of the three costs.

For SGM's unary data term 
we use negated and truncated normalized cross correlation (NCC):
\begin{equation}
C_\pp(d) = 1 - \max(0, \text{NCC}(\pp,d)),
\label{eq:tncc}
\end{equation}
where $\text{NCC}(\pp,d)$ compares $5\!\times\!5$ grayscale image patches
centered at $\pp$ and $\pp-(d,0)^T$ in the left and right image,
respectively. Image intensities are in the range $[0,255]$.  We add a
small value $\epsilon \!=\! 1.0$ to the $\text{NCC}$ denominator to
suppress the effect of noise in untextured regions.  We scale
$C_\pp(d)$ by 255 and round it to the nearest integer. We can use
unsigned shorts for SGM's aggregated costs, which reduces the memory
overhead.

We use NCC as matching cost since it is commonly employed in
real-world systems.  As mentioned, our method is orthogonal to the
choice of matching cost.  Below, and in the supplementary materials,
we also evaluate MC-CNN, the state-of-the-art matching cost by Zbontar
and LeCun \cite{Zbontar2016}, and show that it yields similar performance.

For SGM's smoothness penalty 
(Eqn.~\ref{eqn:smoothness})
we use the following settings:
%\begin{equation}
%P_1 = 100, ~~~ P_2 = P_1 \left(1 + \alpha e^{-|\Delta{I}|/\beta} \right),
%\end{equation}
$P_1 \!=\! 100$, $P_2 = P_1 (1 + \alpha e^{-|\Delta{I}|/\beta} )$,
where $\alpha \!=\! 8$, $\beta \!=\! 10$, and $|\Delta{I}|$ is the
absolute intensity difference at neighboring pixels.  Our choice of
$P_2$ favors large disparity jumps at high-contrast image edges.

\begin{table}
{\small
\begin{tabular*}{\columnwidth}{@{~~~~~~}l@{~ -- ~}l}
\multicolumn{2}{@{}l}{No Prior}\\
\hline
SGM     & Baseline method\\[1mm]
\multicolumn{2}{@{}l}{2D Prior (offset image representation)}\\
\hline
SGM-EPi & Estimated segmented planes \\
SGM-GS  & GT surface \\
SGM-GP  & GT surface, planar approximation \\
SGM-GNi & GT normals (fixed-$z$ ``strawman'') \\[1mm]
\multicolumn{2}{@{}l}{3D Prior (offset volume representation)}\\
\hline
SGM-EPv & Estimated overlapping planes \\
SGM-GNv & GT normals (accurate version) \\
SGM-MW  & Manhattan-world prior \\
\end{tabular*}
}
\caption{The algorithm variants compared in our experiments.  See
Section~\ref{sec:variants} for details.}
\label{tab:variants}
\end{table}

\begin{figure}[t]
\newcommand{\ww}{.75in}
\centering
\begin{tabular}{@{}c@{~~}c@{~~}c@{~~}c@{}}
\includegraphics[width=\ww]{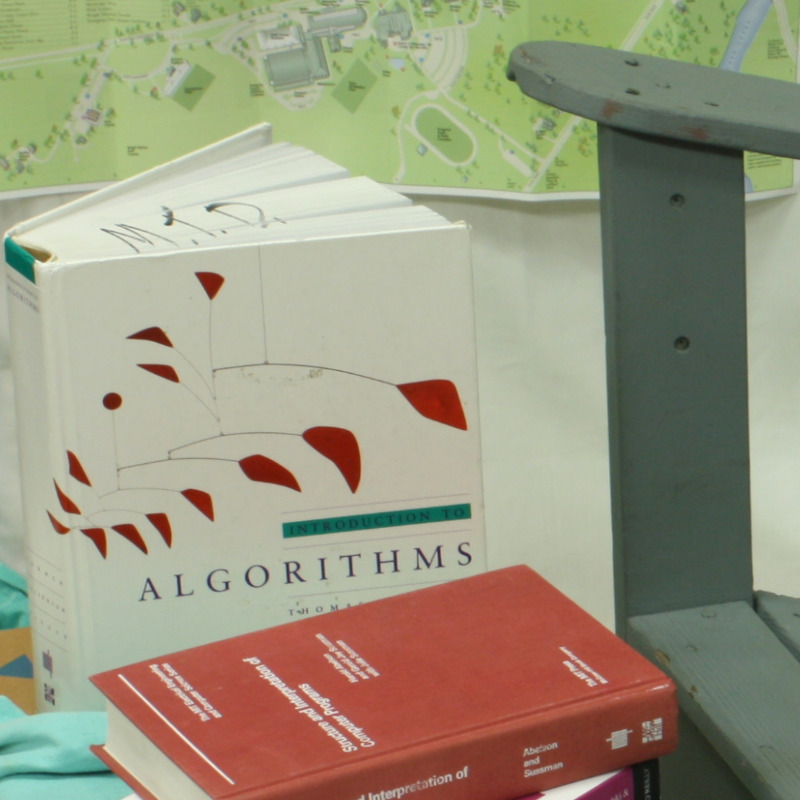} &
\includegraphics[width=\ww]{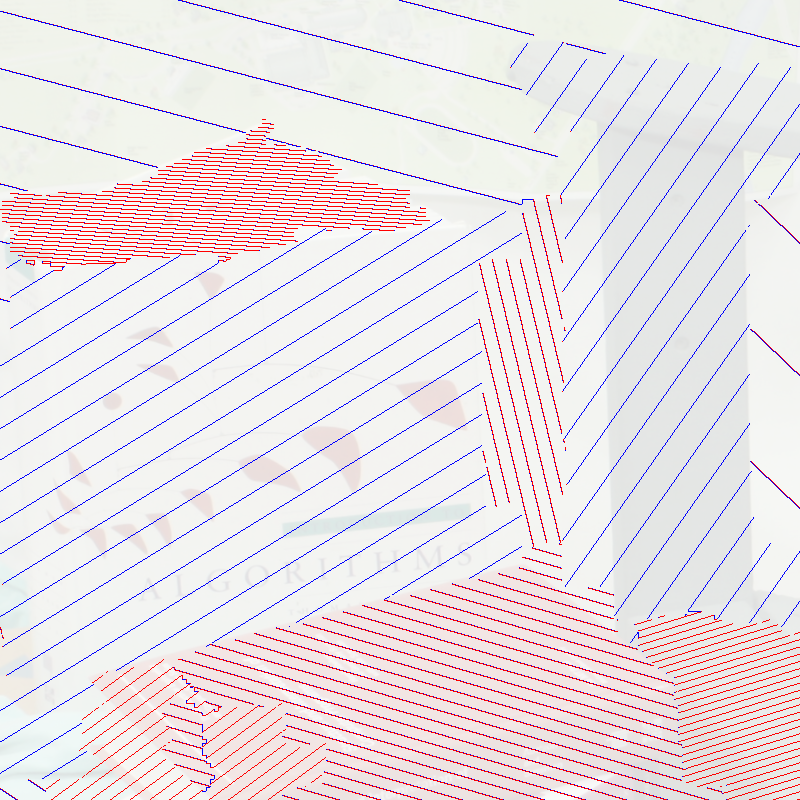} &
\includegraphics[width=\ww]{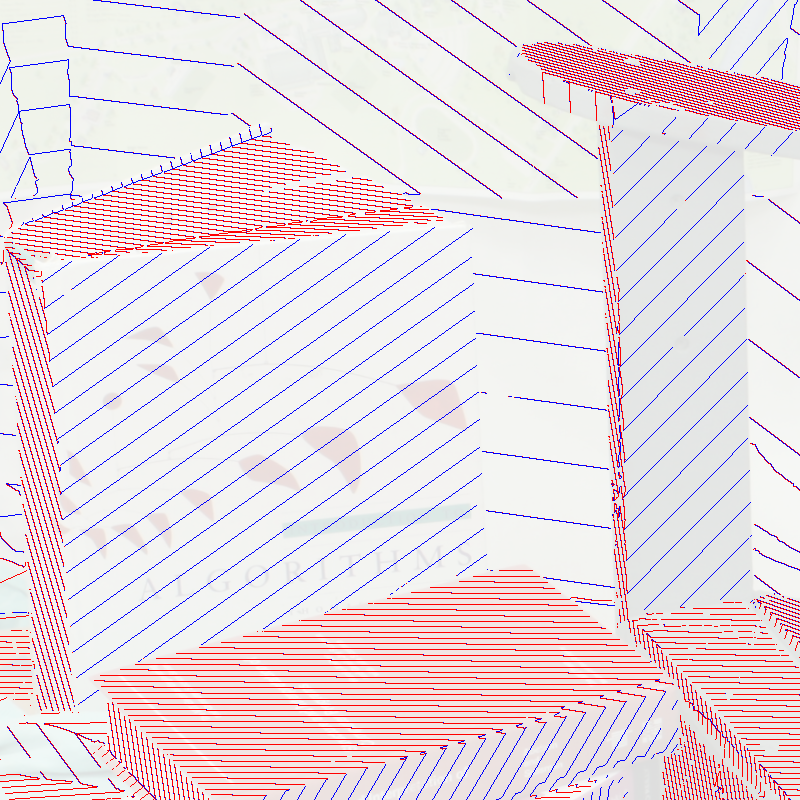} &
\includegraphics[width=\ww]{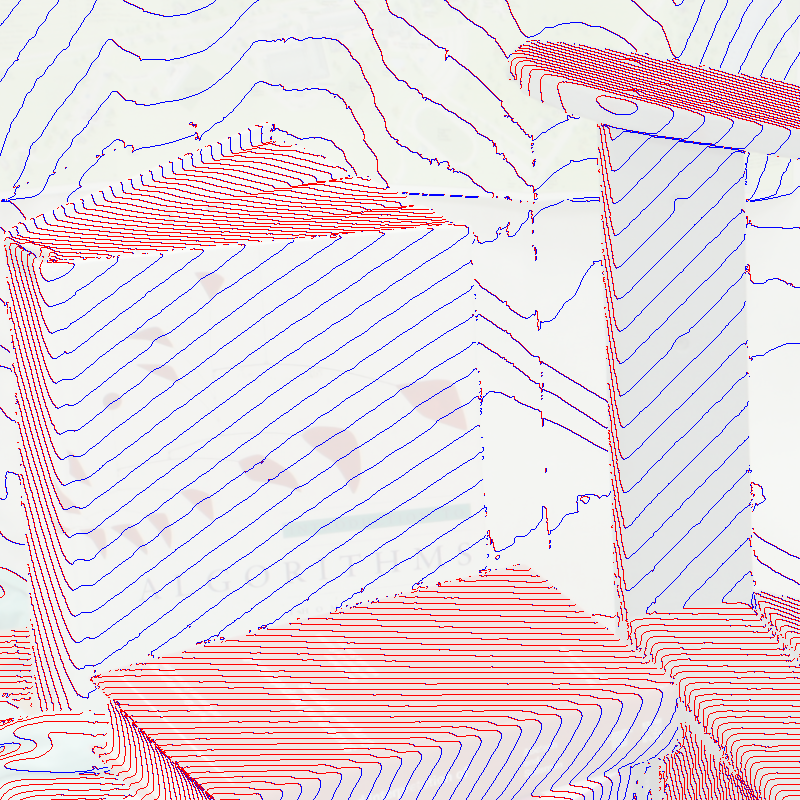} \\
\includegraphics[width=\ww]{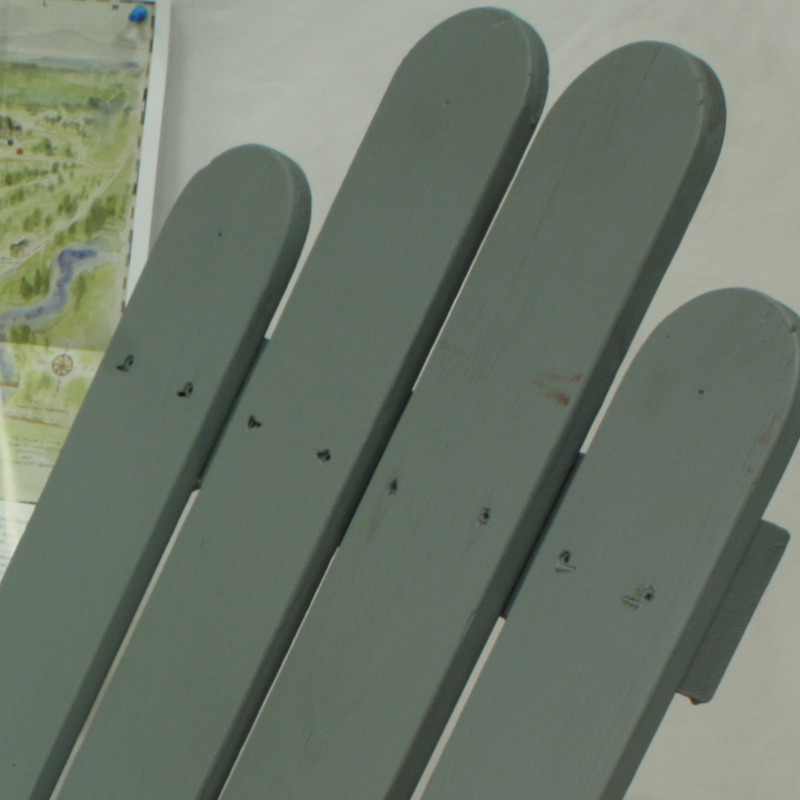} &
\includegraphics[width=\ww]{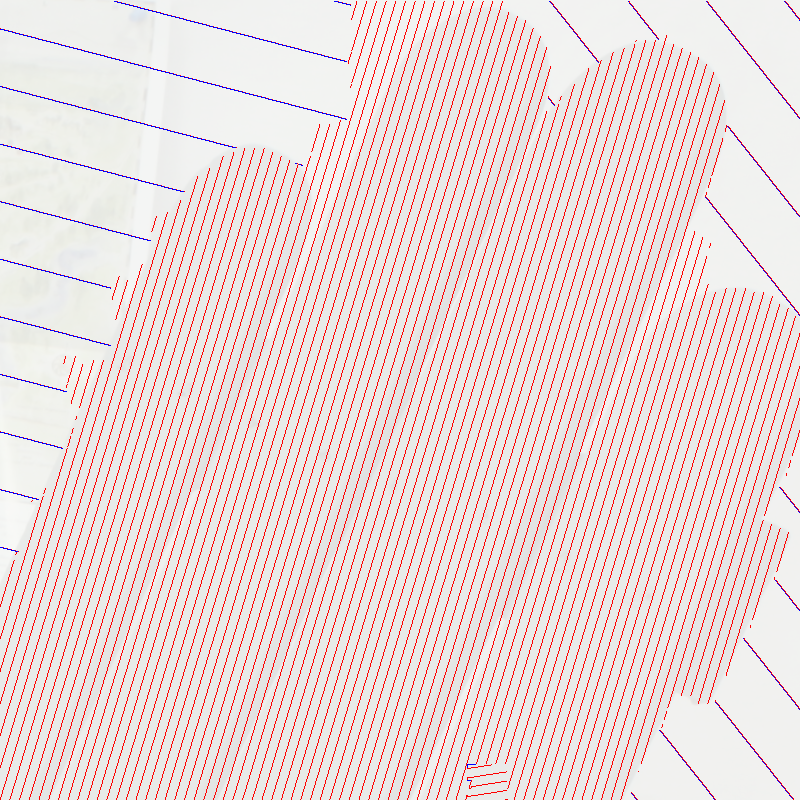} &
\includegraphics[width=\ww]{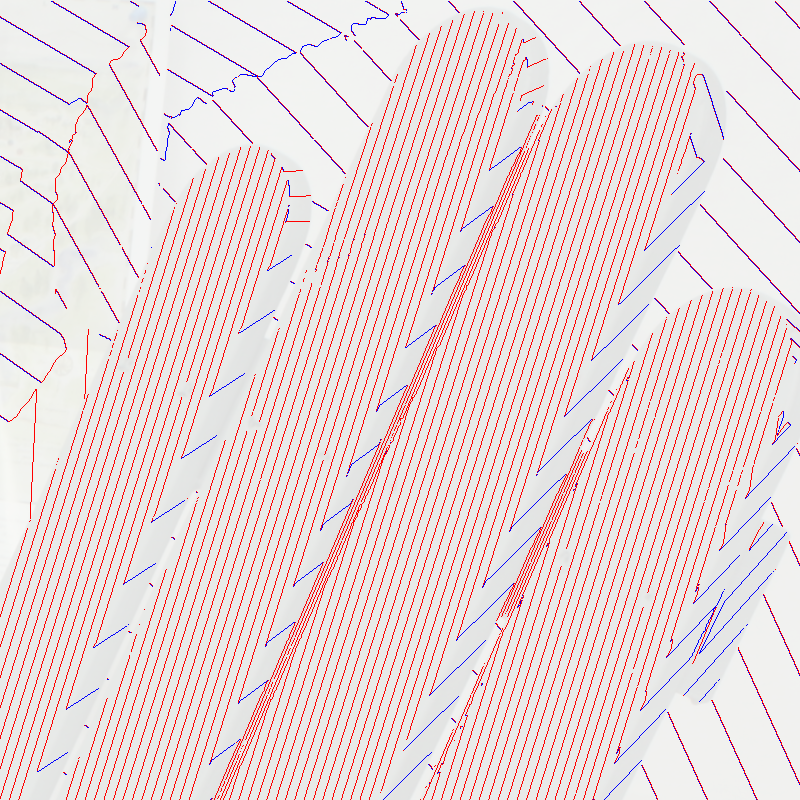} &
\includegraphics[width=\ww]{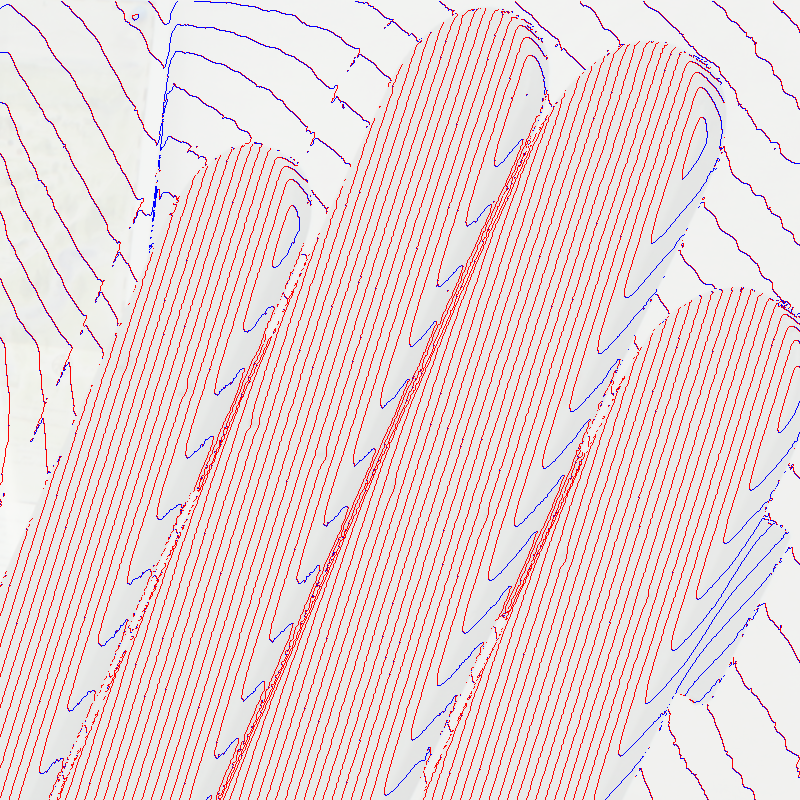} \\
\includegraphics[width=\ww]{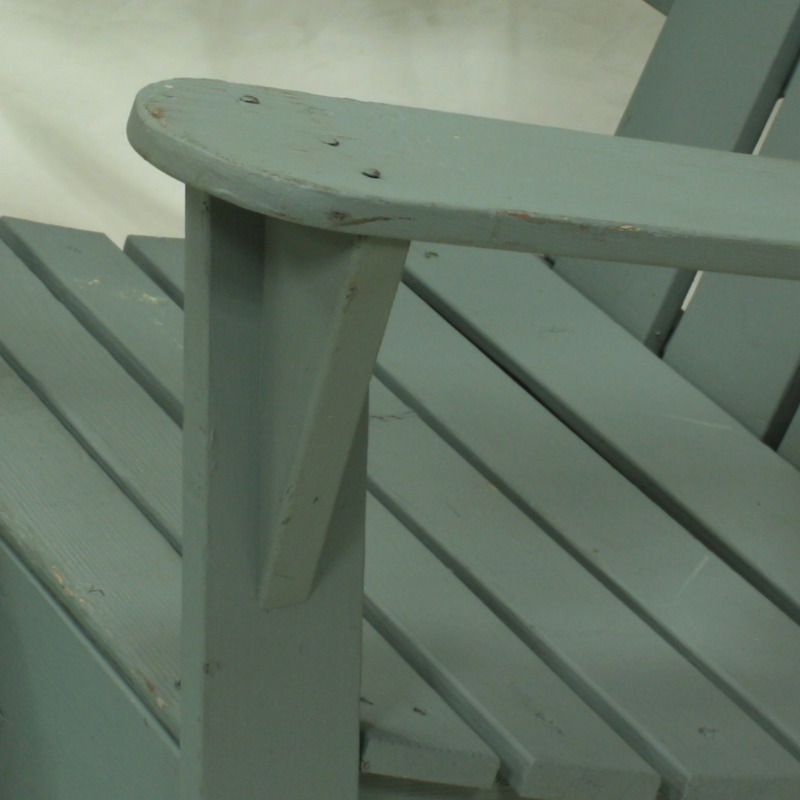} &
\includegraphics[width=\ww]{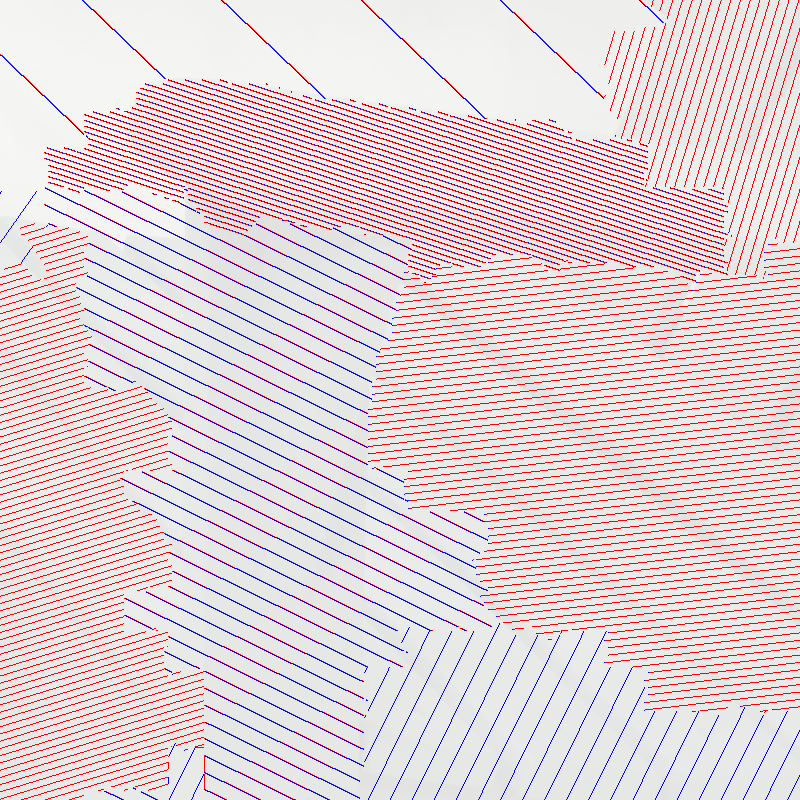} &
\includegraphics[width=\ww]{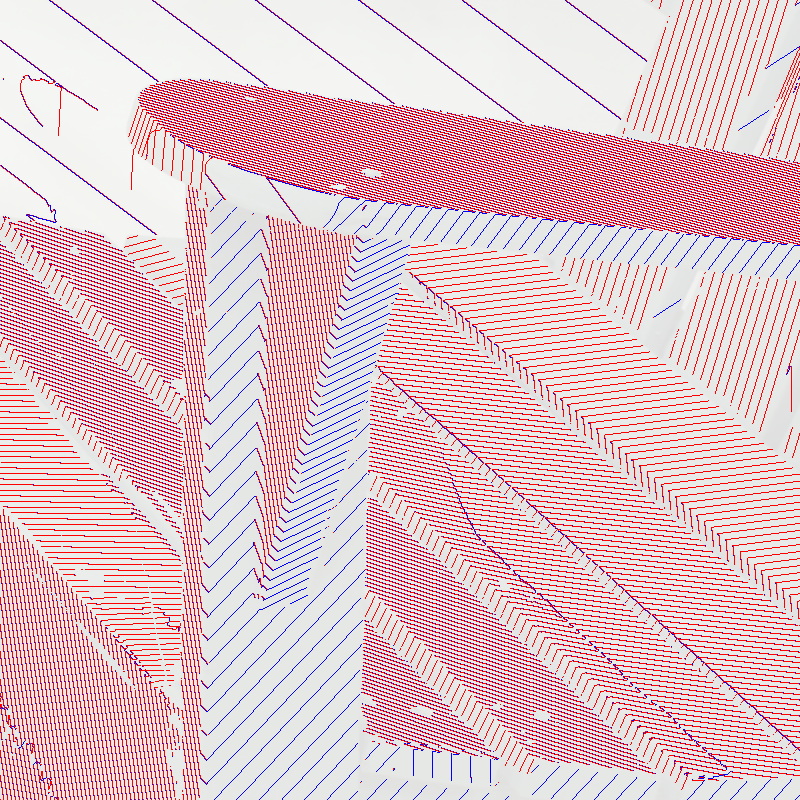} &
\includegraphics[width=\ww]{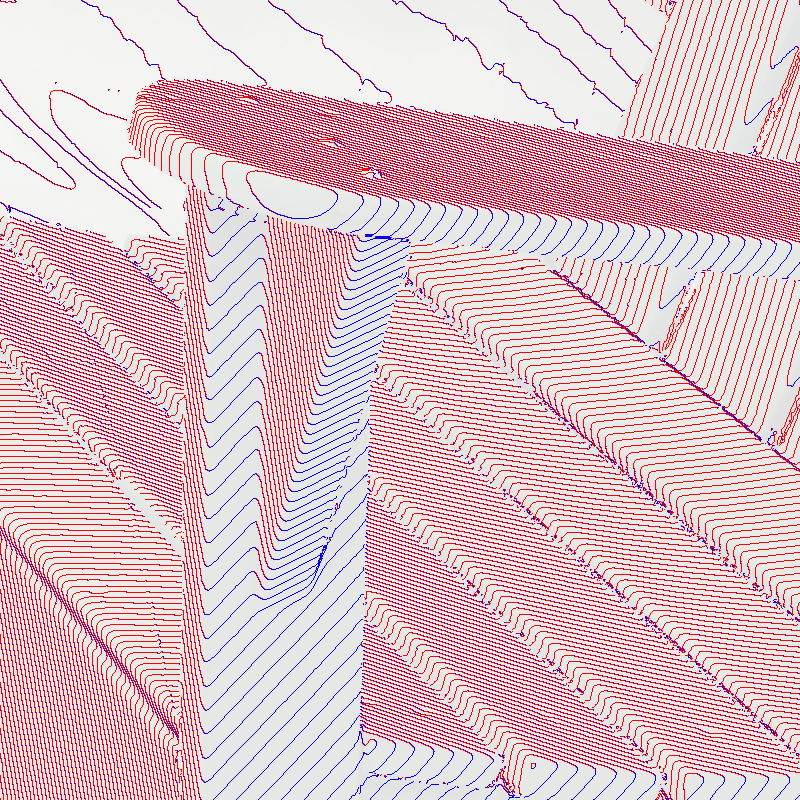} \\[-1mm]
\footnotesize Input &
\footnotesize SGM-EPi &
\footnotesize SGM-GP &
\footnotesize SGM-GS \\[1mm]
\end{tabular}
\caption{Visualization of different 2D orientation priors 
(``offset images'') on zoomed regions of the Adirondack image pair.
%%for three types of priors (see Table~\ref{tab:variants})
}
\label{fig:2dprior}
\end{figure}

\subsection{Algorithm variants}
\label{sec:variants}

In order to evaluate the full potential of SGM-P, we evaluate a number
of different priors.  We will substitute P with a combination of
letters to distinguish the algorithm variants
(see Table~\ref{tab:variants} for a summary 
and Fig.~\ref{fig:2dprior} for visualizations).
The first letter distinguishes priors {G} derived from ground-truth
disparities with priors {E} estimated from the input images.  The
former versions can be considered \emph{oracles} that provide 
an upper bound on the potential benefit of our
idea, while the latter versions give an indication of the actual
realizable benefit.
The second letter denotes the type of surfaces acting as priors: {S}
for arbitrary (e.g., curved) surfaces, {P} for planar surfaces, and
{N} for the case when only surface normals are available.
Finally, we use {i} and {v} to distinguish between 2D priors
%(Section~\ref{sec:2Dprior}) 
only requiring an offset image, and 3D priors 
%(Section~\ref{sec:3Dprior})
requiring an offset volume.

\begin{figure}
\newcommand{\hh}{13.5mm}
\centering
\begin{tabular}{@{}c@{~}c@{~}c@{~}c@{}}
\includegraphics[height=\hh]{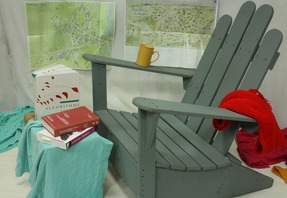} &
\includegraphics[height=\hh]{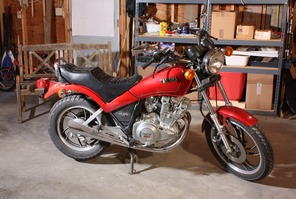} &
\includegraphics[height=\hh]{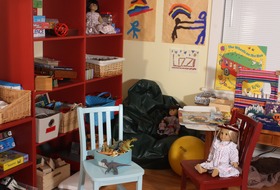} &
\includegraphics[height=\hh]{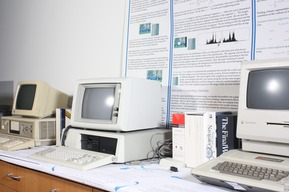} \\[-.5mm]
\includegraphics[height=\hh]{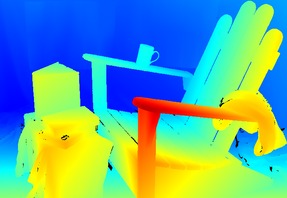} &
\includegraphics[height=\hh]{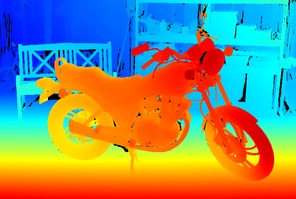} &
\includegraphics[height=\hh]{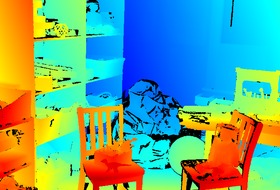} &
\includegraphics[height=\hh]{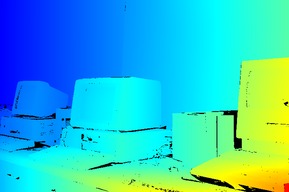} \\[-1mm]
\footnotesize Adirondack &
\footnotesize Motorcycle &
\footnotesize Playroom &
\footnotesize Vintage \\
\end{tabular}
\caption{Challenging high-resolution Middlebury datasets with 
untextured slanted surfaces.}
\label{fig:dataset}
\end{figure}

For the quantitative analysis
we focus on only one {E} variant that estimates priors
from the input images.  We do this by running SGM at a coarser
resolution and clustering the resulting disparities into disparity
plane hypotheses~\cite{Kowdle2012}.  We use these hypotheses and the
associated pixel-to-plane label map to generate 2D and 3D orientation
priors (EPi and EPv).  For the 2D variant, SGM-EPi, we segment the
image into superpixels \cite{Achanta2012}
and then select for each superpixel the plane most often
assigned to its constituent pixels. Superpixels
with low support for any of the plane hypotheses 
%not assigned to any clusters 
are set to an arbitrary fronto-parallel plane.

The more powerful 3D variant, SGM-EPv, allows modeling of multiple
disparity surface hypotheses at the same pixel. We use the same pixel-to-plane
label map as for SGM-EPi but obtain potentially overlapping disparity hypotheses by
bounding each 3D disparity plane by the convex hull of its constituent pixels
in the label map.

For the oracle priors G based on ground-truth disparities we compare all
three surface variants (S, P, and N) in order to explore the benefits and
limitations of the different types of priors.  Of these, SGM-GS uses the
ground-truth disparity surface as 2D prior directly, which is the best
possible prior available.  Next, SGM-GP uses a piecewise-planar approximation of
the ground-truth surface, again as 2D prior, constructed in the same
manner as SGM-EPi.  We omit the suffix ``i'' in both cases since
we do not have corresponding 3D priors. (While
a 3D variant of SGM-GP with multiple overlapping planar hypotheses is possible, we
found that it yielded no benefit over the 2D version.)
Next, SGM-GNv discards the original ground-truth surface and uses
only its normal map, which results in a 3D prior as explained in
Section~\ref{sec:normals}.  
%To investigate whether this 3D prior is indeed necessary,
We also include a ``strawman'' 2D version, SGM-GNi, which we obtain by integrating
a single $z$-surface at an arbitrary depth.
% and converting it into a 2D disparity prior.
Finally, we investigate a 3D normal prior estimated from the images
using a Manhattan-world assumption \cite{Lee2009};
deviating from our naming scheme we simply call it SGM-MW
(more details on this below).

\subsection{Quantitative analysis}

We start by evaluating the promise of SGM-P on a subset of the
high-resolution stereo pairs from the training set of the
Middlebury stereo evaluation v3 \cite{Scharstein2014}, for which
high-quality ground-truth disparities are available.  We select 4 challenging
image pairs with untextured slanted surfaces,
depicted in Fig.~\ref{fig:dataset}.  In the experiments below, we use the
full-resolution (5--6 MP) versions of these datasets;
see the supplementary materials for additional results,
including other resolutions.

\begin{figure}
\centering
\includegraphics[width=\columnwidth]{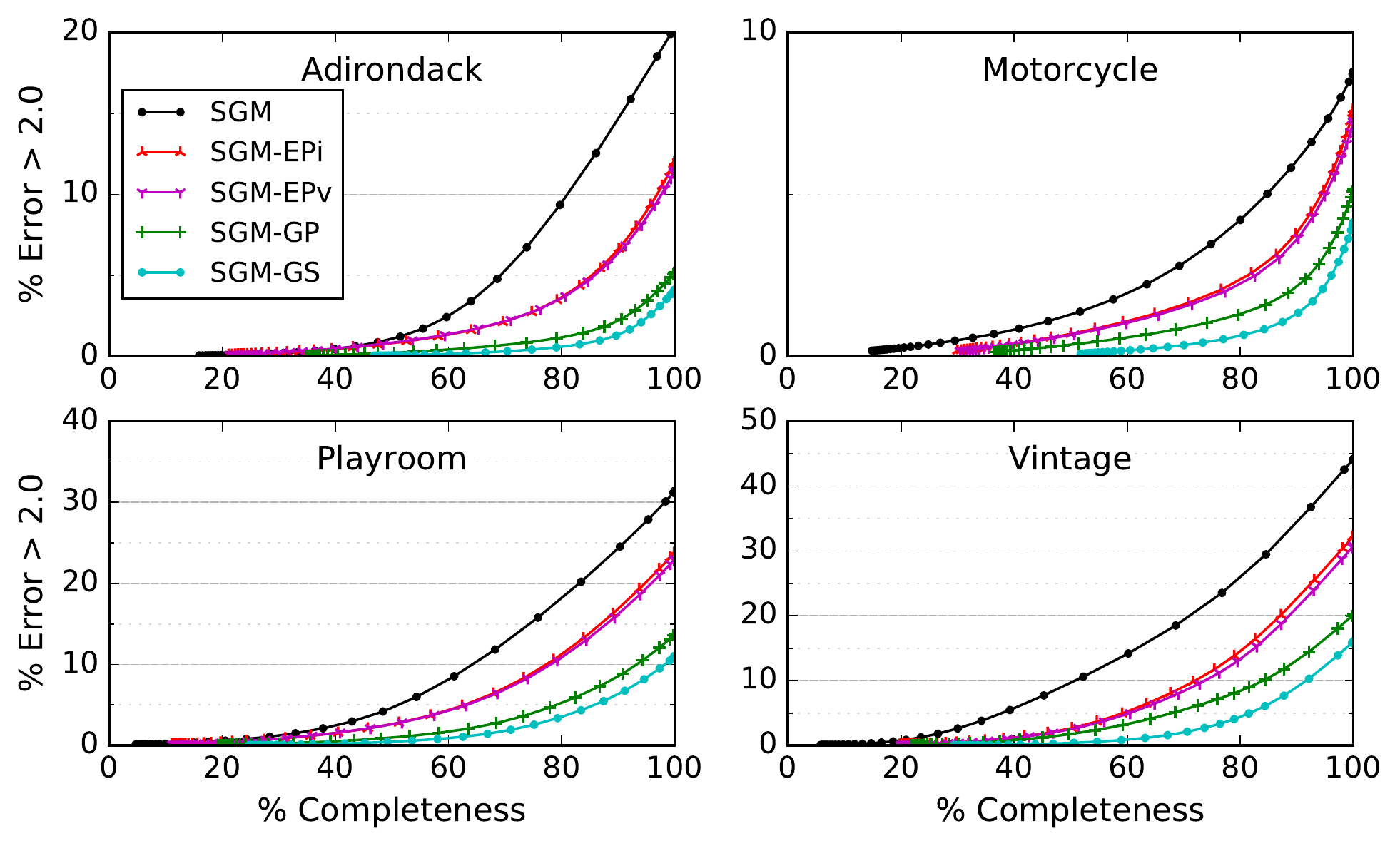}
\caption{ROC curves plotting error rate vs. completeness for baseline SGM,
estimated planar priors SGM-EPi and SGM-EPv, and ground-truth plane and 
surface priors SGM-GP and SGM-GS.}
\label{fig:plots1}
\end{figure}

\begin{figure*}
\newcommand{\hh}{13.2mm}
{\footnotesize
\centering
\begin{tabular}{l@{~~}c@{~}c@{~~}c@{~}c@{~~}c@{~}c@{~~}c@{~}c@{}}
&
\multicolumn{2}{c}{Adirondack} &
\multicolumn{2}{c}{Motorcycle} &
\multicolumn{2}{c}{Playroom} &
%\multicolumn{2}{c}{Recycle} &
\multicolumn{2}{c}{Vintage} \\
&
disparities & error map &
disparities & error map &
disparities & error map &
disparities & error map \\
\shortstack{\footnotesize SGM\\[4mm] \mbox{ }} &
\includegraphics[height=\hh]{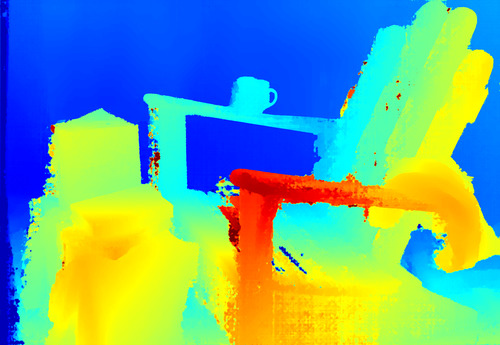} &
\includegraphics[height=\hh]{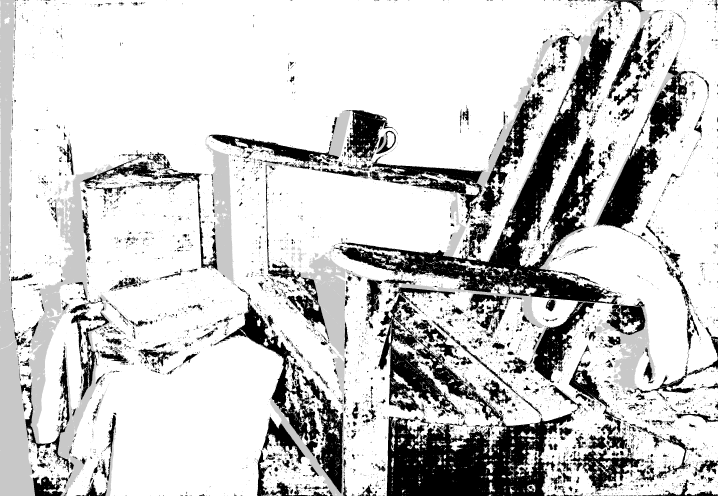} &
\includegraphics[height=\hh]{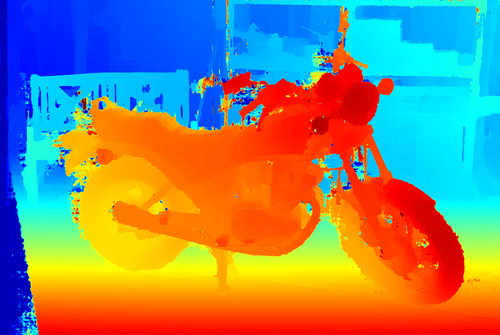} &
\includegraphics[height=\hh]{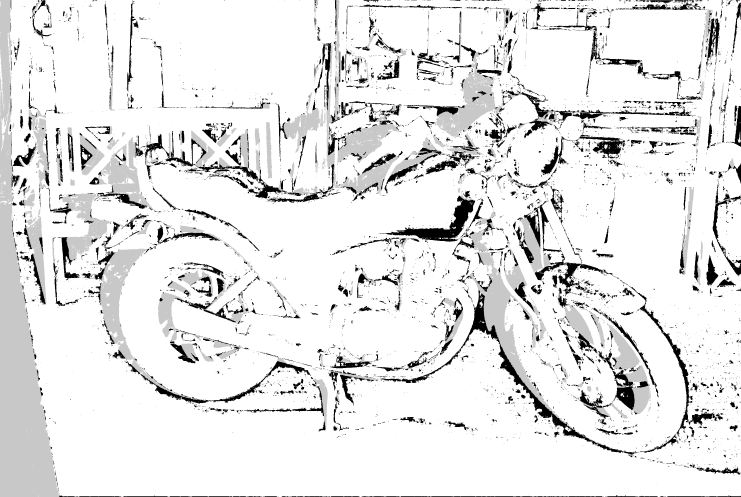} &
\includegraphics[height=\hh]{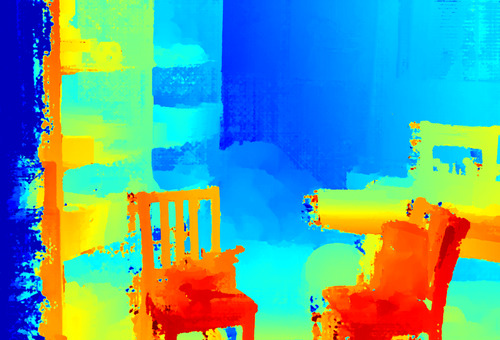} &
\includegraphics[height=\hh]{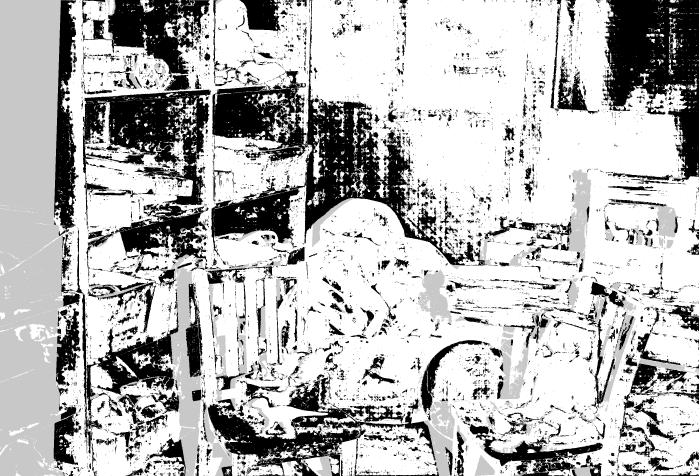} &
\includegraphics[height=\hh]{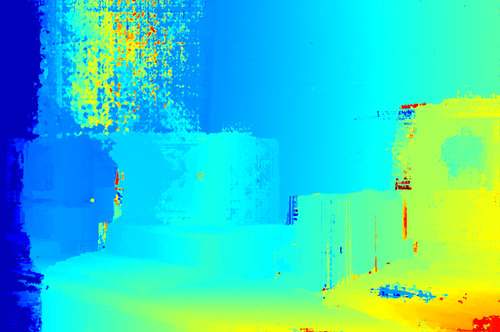} &
\includegraphics[height=\hh]{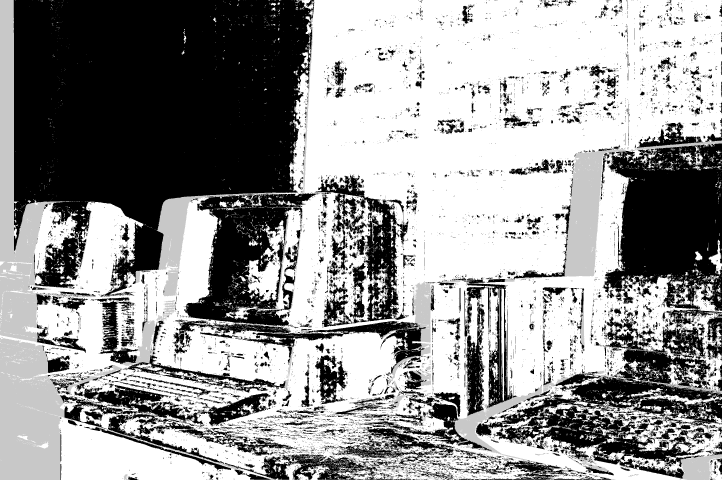} \\[-.5mm]
\shortstack{\footnotesize SGM-\\EPi\\[2mm] \mbox{ }} &
\includegraphics[height=\hh]{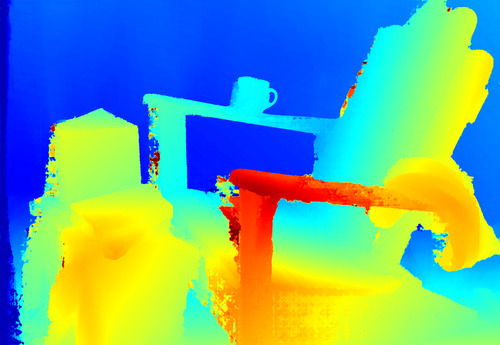} &
\includegraphics[height=\hh]{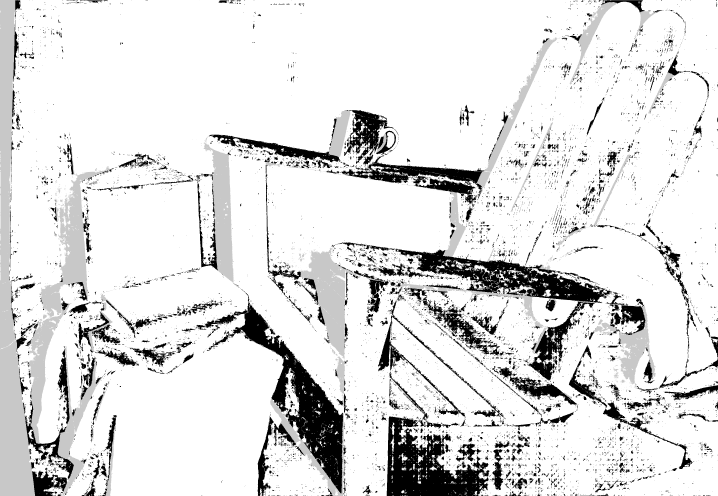} &
\includegraphics[height=\hh]{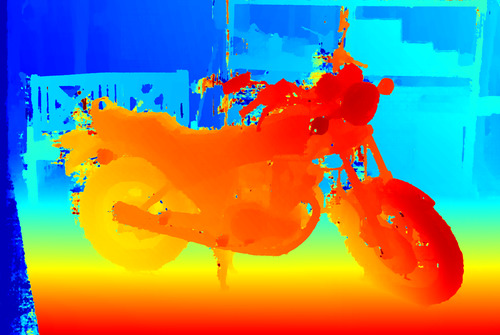} &
\includegraphics[height=\hh]{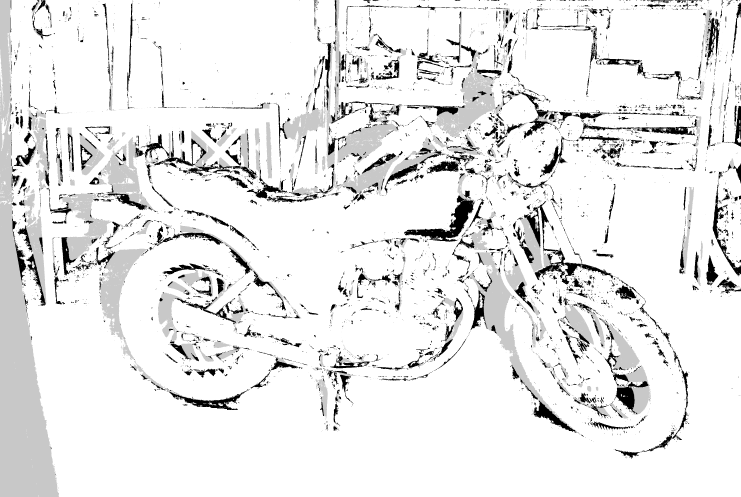} &
\includegraphics[height=\hh]{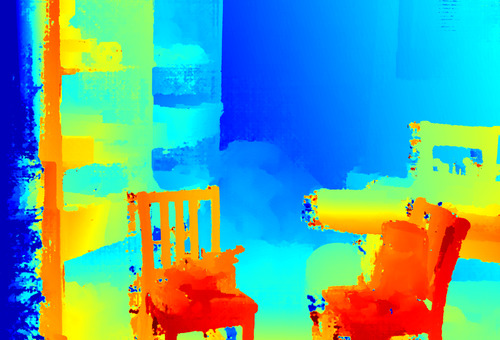} &
\includegraphics[height=\hh]{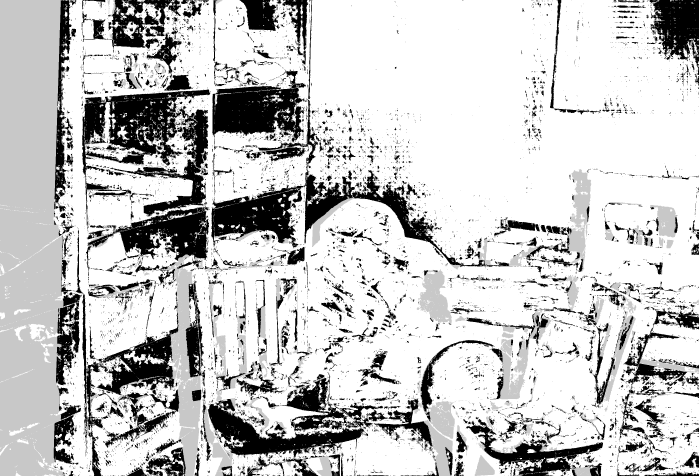} &
\includegraphics[height=\hh]{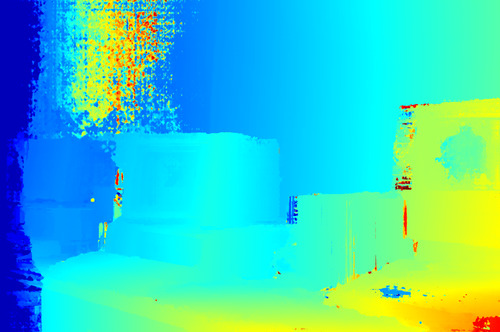} &
\includegraphics[height=\hh]{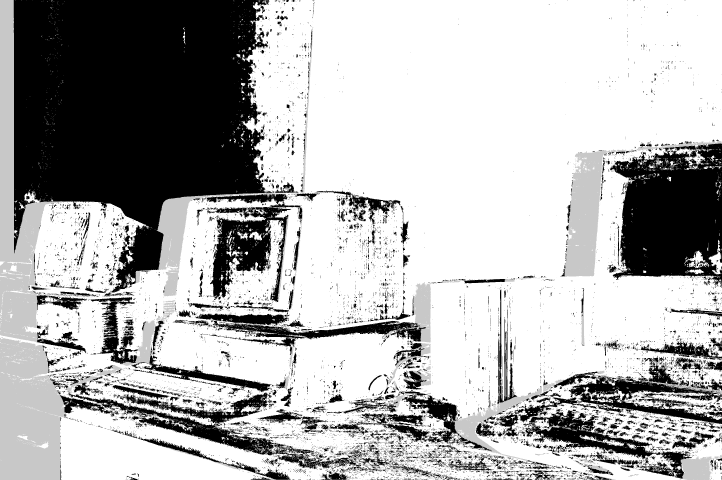} \\[-.5mm]
\shortstack{\footnotesize SGM-\\GS\\[2mm] \mbox{ }} &
\includegraphics[height=\hh]{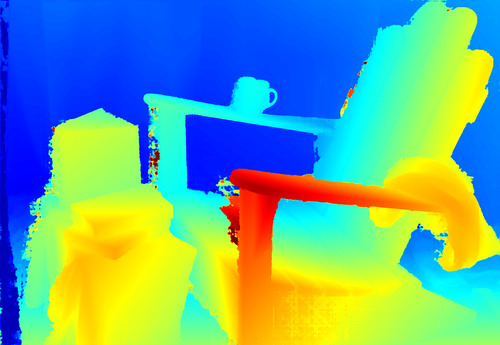} &
\includegraphics[height=\hh]{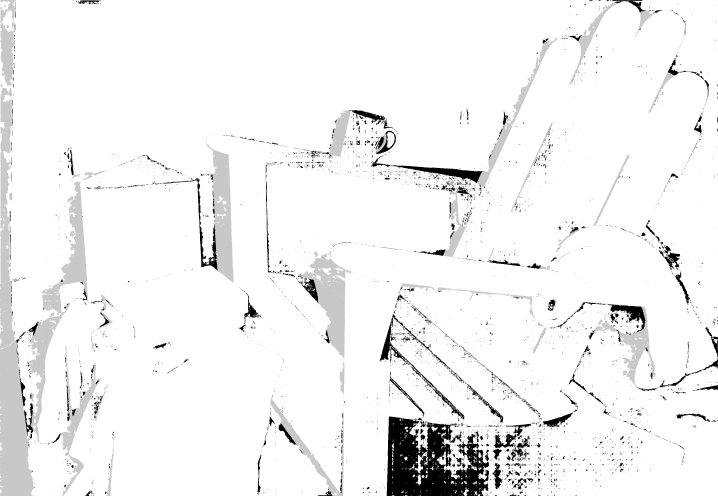} &
\includegraphics[height=\hh]{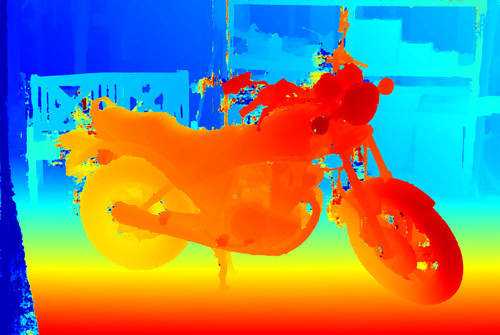} &
\includegraphics[height=\hh]{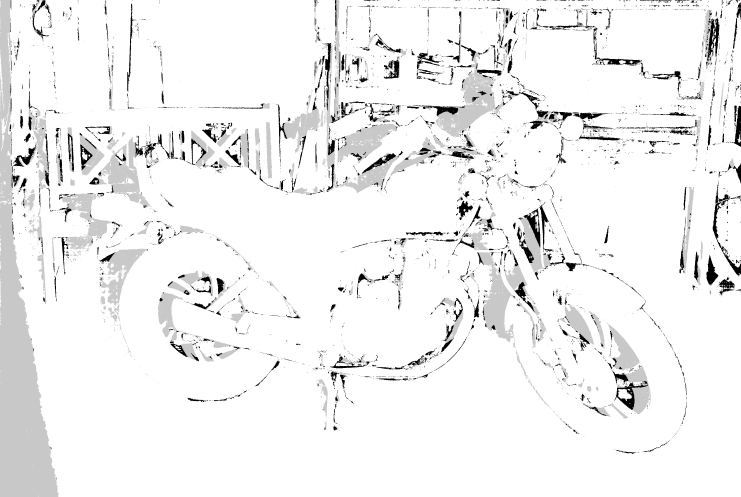} &
\includegraphics[height=\hh]{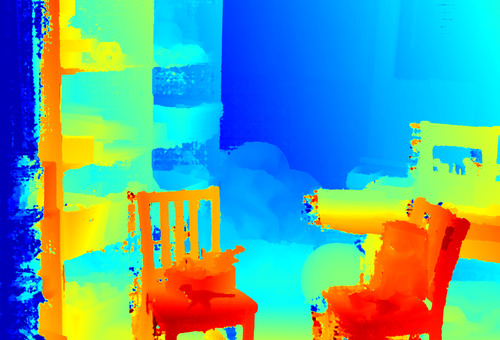} &
\includegraphics[height=\hh]{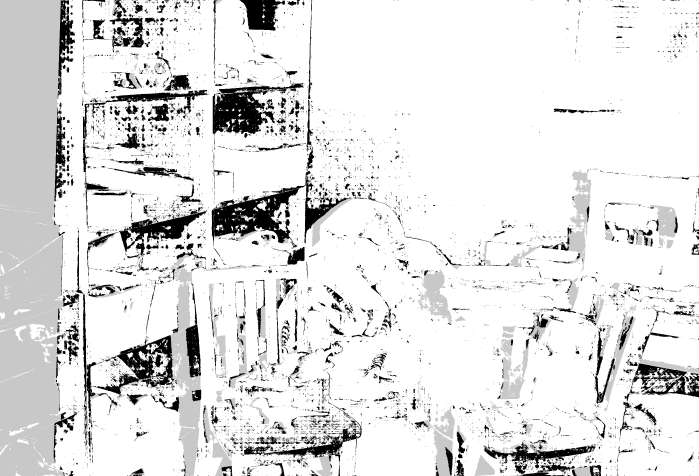} &
\includegraphics[height=\hh]{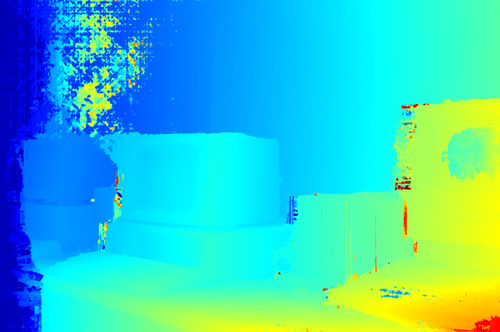} &
\includegraphics[height=\hh]{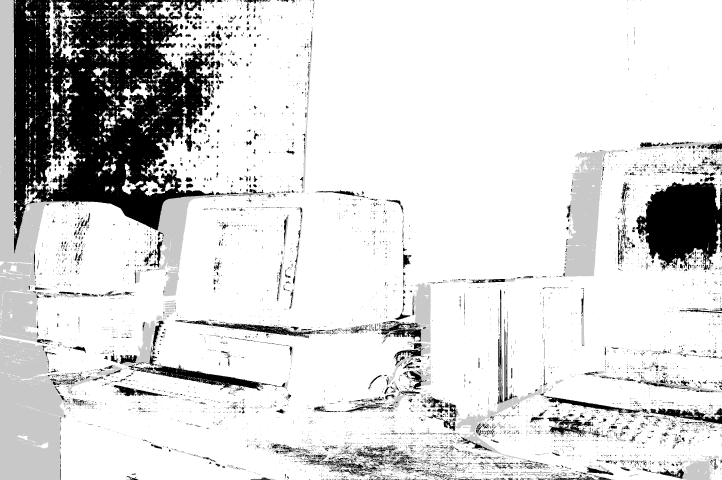} \\
\end{tabular}
}
\caption{Disparity maps and error maps corresponding to the plots
in Fig.~\ref{fig:plots1} at 100\% completeness.  Black regions in the
error maps indicate disparity errors $>$ 2.0 in non-occluded regions.
SGM-EPi and SGM-GS yield a noticeable reduction of errors on slanted
surfaces.}
\label{fig:results1}
\end{figure*}

%%\subsubsection{Promise of SGM-P}

In our first experiment, we compare the baseline SGM method with our
SGM-P algorithm using both estimated planar priors (SGM-EPi and SGM-EPv) and
ground-truth priors (SGM-GP and SGM-GS).  Fig.~\ref{fig:plots1} shows
disparity error rates (percentage of pixels whose disparity error is
greater than $t$=2.0) as a function of completeness (inverse
sparsity).  We obtain disparity maps with increasing completeness by
raising the allowable uncertainty $U_\pp$ (Eqn.~\ref{eqn:uncertainty})
from 0 to $U_{\max}$. Our plots are similar to ROC curves and
allow the comparison of sparse (or semi-dense) stereo methods that
leave uncertain regions unmatched
\cite{kostliva2007}.  The error rate for the dense result (100\% completeness)
is the right-most point on each curve.

The plots in Fig.~\ref{fig:plots1} show that the four variants of our
SGM-P algorithm all significantly outperform the baseline SGM
algorithm on these four image pairs.  As expected, the best
performance is obtained with perfect orientation priors derived from
the ground-truth surface (SGM-GS), which yields a dramatical
improvement over SGM, with error rates ranging from one half to one fifth of
the original errors.  As mentioned, this provides an upper bound on
the potential benefit of our idea.  A more realistic upper bound is
given by SGM-GP, which utilizes piecewise planar priors derived
from the ground-truth disparities.
This results in a slight decrease in performance compared to SGM-GS, but
still a dramatical increase over the baseline.

Most importantly, even without utilizing ground-truth information, we
still get a significant improvement from planar priors
estimated from the input images
(SGM-EPi and SGM-EPv).  The two versions, which are almost indistinguishable
in terms of performance, achieve
between 25\% and 50\% of the upper bounds,
resulting in an improvement over the original SGM errors by 13--41\%.
For now we will focus on the simpler SGM-EPi method; we will discuss
the potential of 3D priors (SGM-EPv) below.
Fig.~\ref{fig:results1} shows the disparity maps and
error maps for the dense results (100\% completeness)
for SGM, SGM-EPi, and SGM-GS.

\begin{figure}
\centering
\includegraphics[width=\columnwidth]{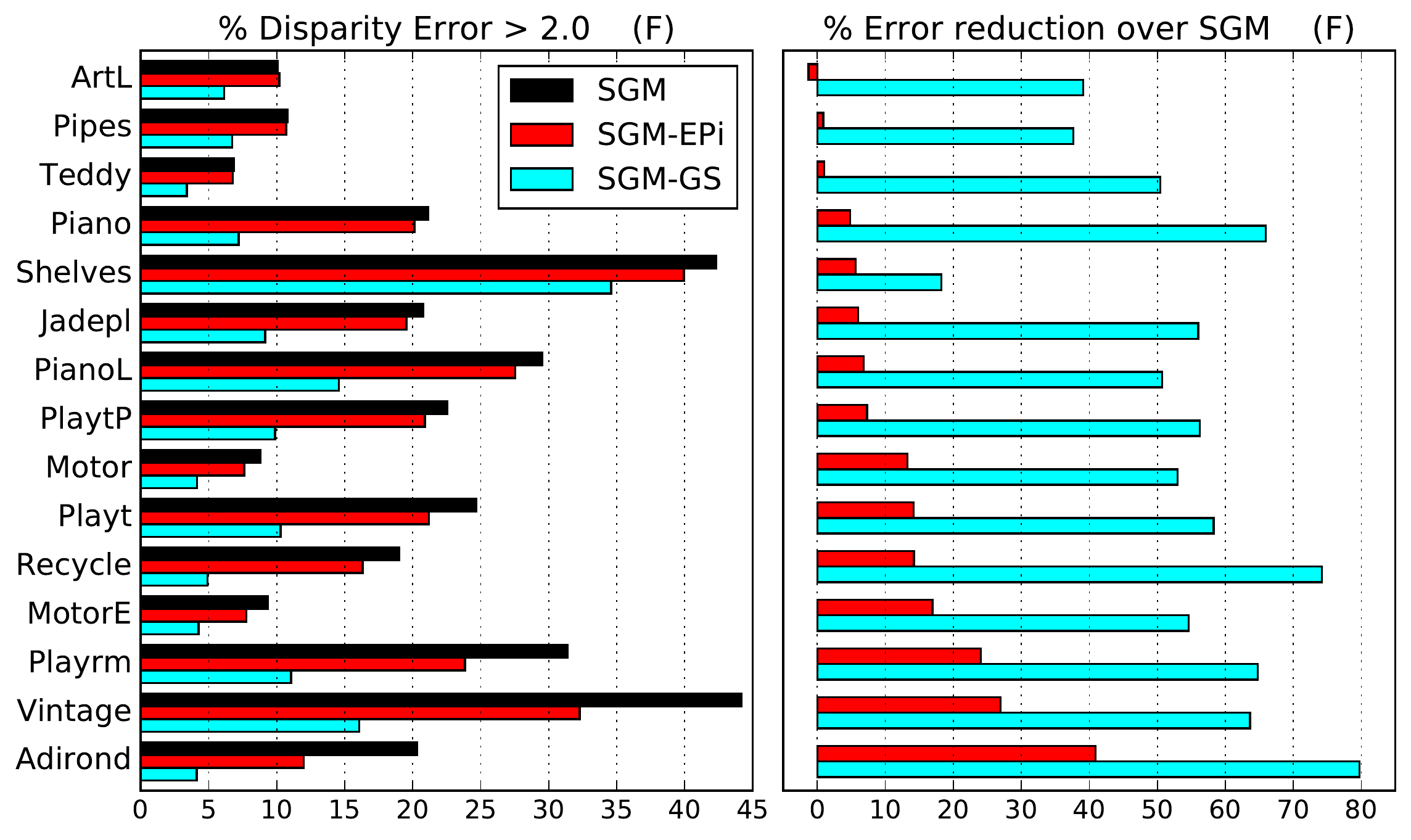}
\caption{Performance on all 15 Middlebury training sets at full resolution,
sorted by increasing performance gain (error reduction) of SGM-EPi over SGM.}
\label{fig:barplot}
\end{figure}

It should be noted that the 
% potential 
benefit of SGM-P strongly
depends on the scene structure.  In scenes with mostly fronto-parallel
surfaces, SGM-P
%, even with ground-truth priors, 
yields little
improvement.  An important question is whether estimated priors can
hurt the performance.  Fig.~\ref{fig:barplot} shows the
performance of SGM-EPi and SGM-GS on all 15 Middlebury training pairs,
sorted by rate of error reduction of SGM-EPi over SGM.
It can be seen that the performance gains range 
from around  -1\% to 41\%, with an average gain of 12\%.
Importantly, the performance never significantly decreases.
We see the same trend for other matching costs; see the
supplementary materials.

\begin{table}
\centering
\includegraphics[width=\columnwidth]{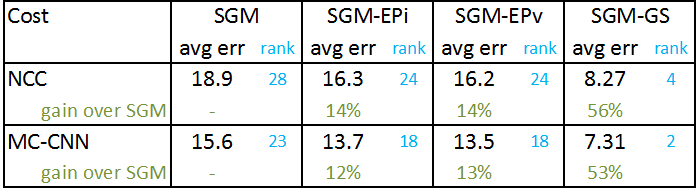}
\caption{Performance of different SGM-P variants and matching costs
on the Middlebury online evaluation for the training sets.}
\label{tab:midd}
\vspace{-3mm}
\end{table}

We also submitted the results of SGM-EPi to the Middlebury stereo
evaluation \cite{MIDDLEBURY-WEB}.  Since only a single submission per
paper is allowed, we do not have a baseline for the 15 Middlebury test
pairs.  We thus cannot show the improvement ratios for the test pairs,
as we do in Fig.~\ref{fig:barplot} for the training pairs.  The public %%online
evaluation results, however, show that our method outperforms all
existing SGM entries in the public table (especially the
full-resolution SGM entry). The largest performance gains are on
scenes with slanted textureless surfaces, including Classroom,
Crusade, and Stairs.  SGM-EPi ranks 20th and 24th overall on test and
training sets, respectively; among the full-resolution submissions it
ranks 3rd on both sets.  Table~\ref{tab:midd} shows the official (weighted)
average training error rates, as well as the table ranks, for the different
SGM-P variants for both NCC and MC-CNN \cite{Zbontar2016} matching costs.
While MC-CNN yields slightly lower errors, both costs result in
similar performance gains.
Recall that our goal is \emph{not} to create a top-ranked stereo
method, but rather to improve upon SGM, one of the most widely-used
stereo methods.  The rankings clearly show the potential of SGM-P for
high-resolution stereo matching.

Finally, the fact that SGM-EPi and SGM-EPv produce very similar numerical
results is not too surprising since most regions in the Middlebury
images can be well explained with single surfaces.
%, and we also do not have a good mechanism for generating multiple
%proposals.
In the supplementary materials we show qualitative evidence that
SGM-EPv is better at recovering surface creases by utilizing multiple 
overlapping hypotheses.  
Harnessing the full power of SGM-EPv, however, would require
more powerful methods for generating priors that extend over larger
regions of the image.

%\subsubsection{2D vs. 3D priors}

\ignore{ OLD ==================================================
We now turn to the comparison between the
2D and 3D versions of the estimated planar priors,
SGM-EPi and SGM-EPv.  On the Middlebury images, these two version
produce very similar numerical results.  This is not too
surprising.  Recall that SGM-EPv allows modeling of different surface
slants at different depths.  However, our current prior estimation
method, which fits planes to matching results at a lower resolution,
is not particularly good at generating multiple proposals.  In
addition, most regions in the Middlebury images can be well explained
with single surfaces.  To harness the full power of SGM-EPv we need
more powerful methods for generating priors that extend over larger
regions of the image.
%which would help for instance to reconstruct fence-like structures
%at different slants.
%

However, we have found qualitative examples where 3D priors beat 2D priors
(see the supplementary materials for images).
On challenging indoor stereo pairs SGM-EPv tends to produce smoother
results near discontinuities and at creases between planes.
Since SGM-EPi only has a single orientation prior per pixel, its
performance degrades when the pixel-to-plane labeling is noisy and
incomplete. SGM-EPv allows multiple disparity hypotheses and is thus
more robust in the presence of noisy labels.
=============================================}

\begin{figure*}[t]
\newcommand{\ww}{1.25in}
\centering
\begin{tabular}{@{}c@{~~~}c@{~~~}c@{~~~}c@{~~~}c@{}}
\includegraphics[width=\ww]{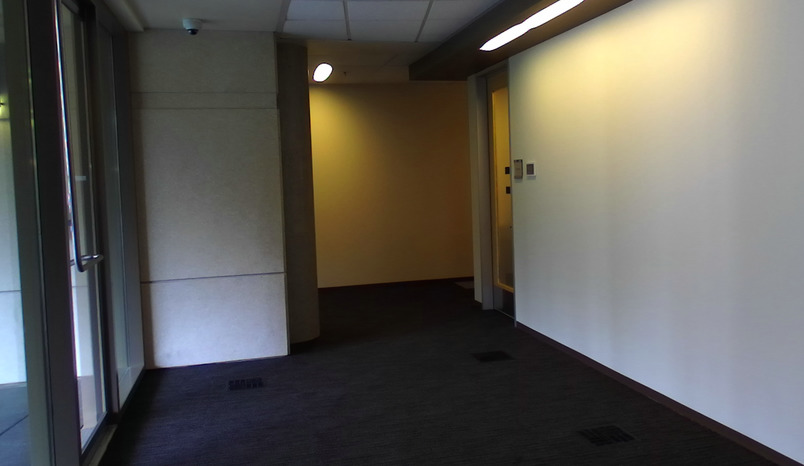} &
\includegraphics[width=\ww]{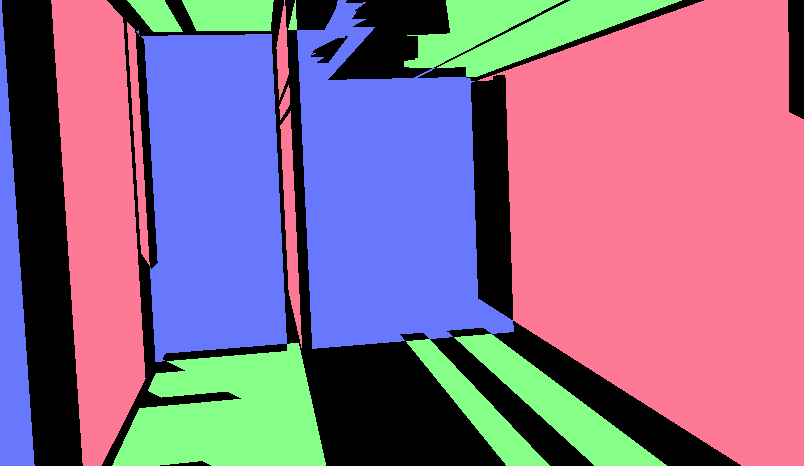} &
\includegraphics[width=\ww]{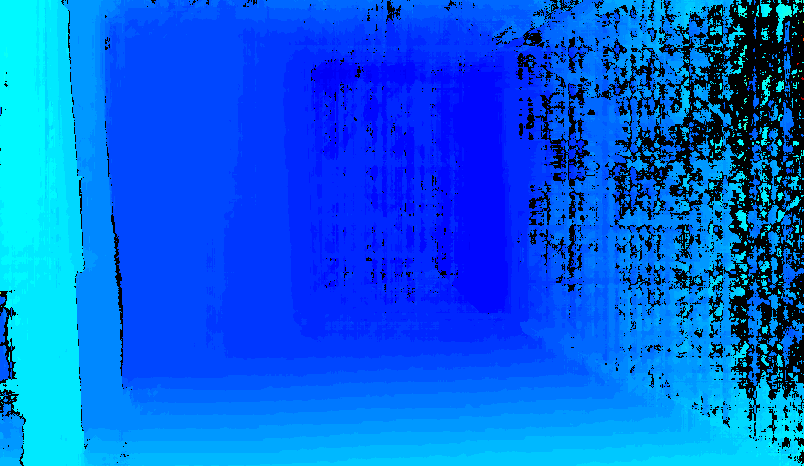} &
\includegraphics[width=\ww]{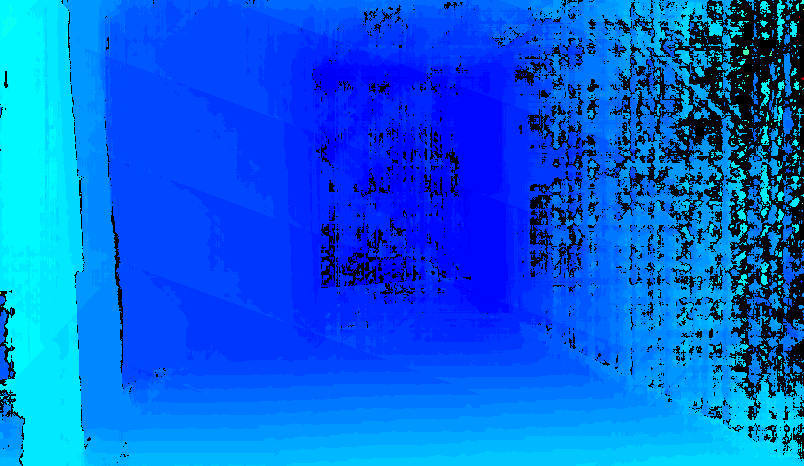} &
\includegraphics[width=\ww]{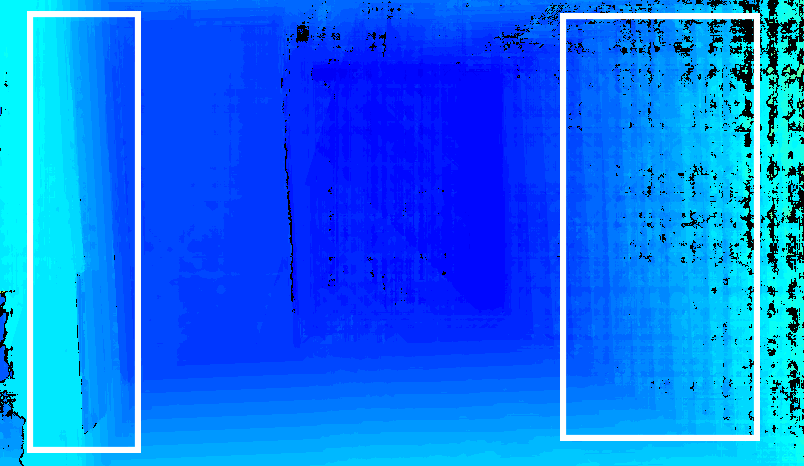} \\
\includegraphics[width=\ww]{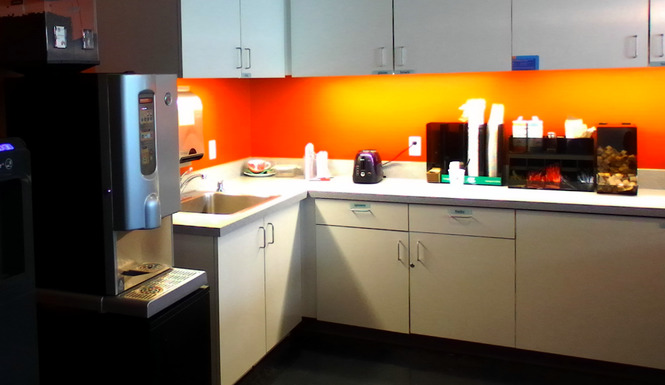} &
\includegraphics[width=\ww]{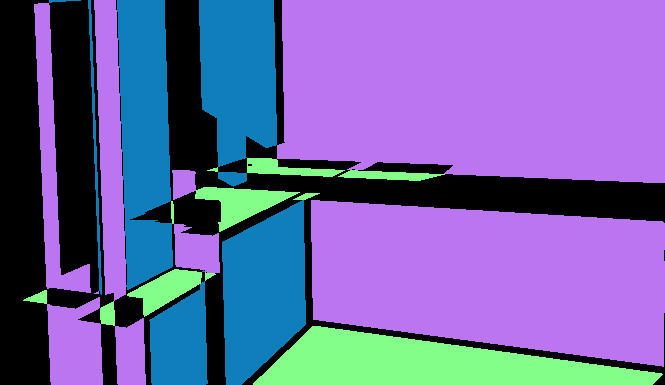} &
\includegraphics[width=\ww]{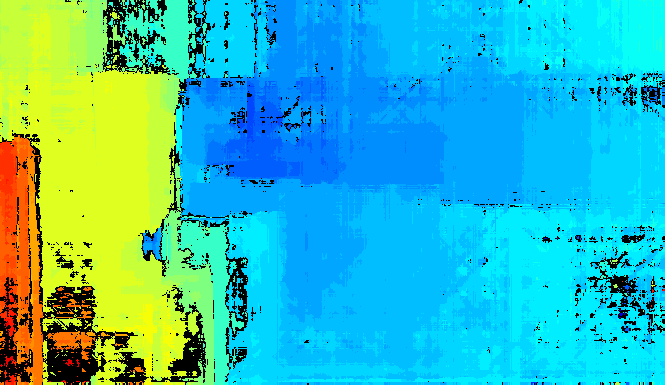} &
\includegraphics[width=\ww]{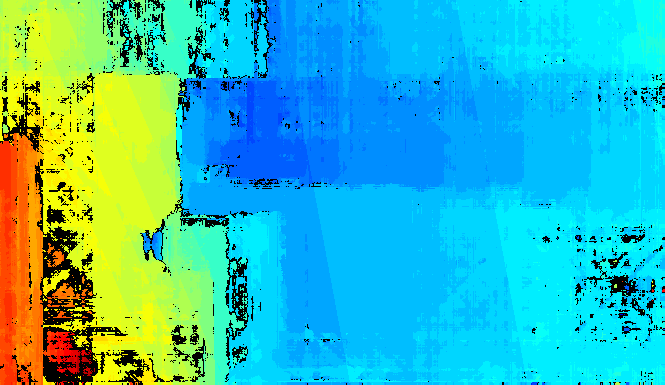} &
\includegraphics[width=\ww]{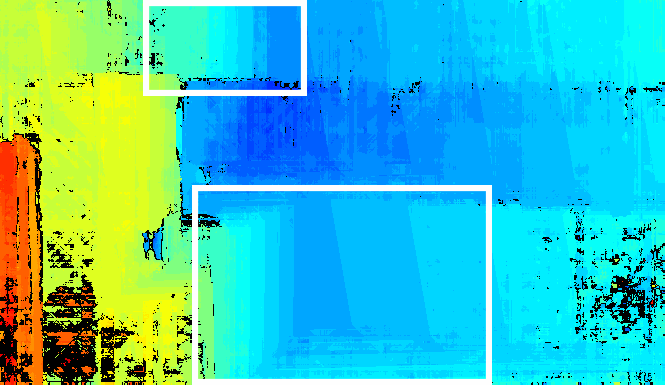} \\[-1mm]
\footnotesize (a) Input image &
\footnotesize (b) Normal map &
\footnotesize (c) SGM &
\footnotesize (d) SGM-EPi &
\footnotesize (e) SGM-MW \\[1mm]
\end{tabular}
\caption{(a, b) Input images and Manhattan-world normal estimates \cite{Lee2009}.
(c, d) Both SGM and SGM-EPi have trouble reconstructing some of the untextured 
slanted surfaces.  (e) SGM-MW utilizes the normal
map (b) and does a better job, in particular in the highlighted regions.  
Note that incorrect or missing normal information
does not hurt performance if sufficient texture is present.}
\label{fig:manhattan}
\end{figure*}

\subsection{Manhattan-world priors}

We exploit Manhattan-world layouts as a type of surface normal priors
in our SGM-MW method. Using a method for scene layout recovery based
on vanishing points \cite{Lee2009}, we obtain a semi-dense pixel
labeling of the Manhattan world's principal surface normals in the
left input image (Fig.~\ref{fig:manhattan}b).  We turn these normal priors
into 3D disparity orientation priors by rendering planar segments in
scene ($z$) space.  Rather than covering the depth range uniformly, we
select depths with local evidence for a surface, found by running SGM
at a coarser resolution. Instead of fitting planes to these
disparities (as we do for SGM-EPi) we convert the disparity map to $z$
space, fit constrained planes with known orientation to the 3D points
and convert those planes back to $d$ space (see
Section~\ref{sec:normals}).  Finally, these disparity planes are
rasterized and used as 3D priors.

Fig.~\ref{fig:manhattan} shows qualitative results of SGM, SGM-EPi,
and SGM-MW on two challenging indoor image pairs. We see that both SGM
and SGM-EPi struggle in regions with slanted untextured surfaces,
where SGM-EPi does not find a supporting plane. However, SGM-MW
recovers smoother slanted surfaces in these regions by utilizing the
Manhattan-world normal estimates.
 
In the supplementary materials we also test our surface normal prior
idea with oracle priors derived from ground-truth data. Recall that a
planar surface patch with known normal $n$ but unknown depth $z$
yields a family of disparity planes whose orientation depends on $z$,
which requires a 3D prior (SGM-GNv).  We show that modeling this
depth dependance is crucial by demonstrating
that SGM-GNv is significantly more accurate than the ``strawman'' 2D
version SGM-GNi, which integrates the surface normals in scene space
at one arbitrary depth and uses the resulting disparity surface as a
2D prior.

\subsection{Runtimes}

%We also tried robust plane fitting to sparse feature matches\cite{sinha-cvpr2014}.
%The estimated disparity planes were sometimes accurate but a dense labelmap had to
%be generated from sparse matches using superpixels and hole-filling but this labelmap
%was in general poorer than the one obtained from coarse stereo matching. Therefore
%we only evaluated our method with coarse stereo based planes.

\begin{figure}
\centering
\includegraphics[width=3.1in]{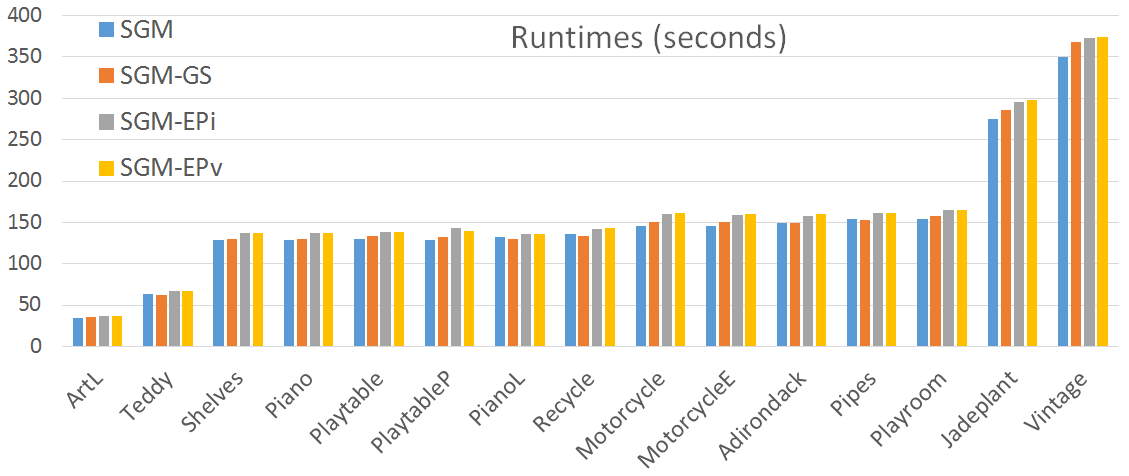}
\caption{Comparison of runtimes.}
% for SGM, SGM-EPi and SGM-GS. 
%SGM-GS has minimal overhead over SGM. 
%SGM-EPi and SGM-EPv have a small additional overhead due to the
%generation of the orientation priors.}
\label{fig:runtime}
\vspace{-3mm}
\end{figure}

Fig.~\ref{fig:runtime} compares the runtime of various SGM-P
versions with the baseline SGM method on the 15 full-resolution
Middlebury training pairs.  All timings were measured on a computer
with a 3.4 GHz Xeon E5-2643 v4 processor and 32 GB RAM.  Our C++
implementation is not yet optimized for speed.  However, the timings
show that SGM-GS has almost no overhead over SGM.  SGM-EPi and SGM-EPv
have similar runtimes, with average runtime overheads over SGM of
about 7\%.  This overhead is mainly due to the cost of extracting the
orientation priors.

\section{Conclusion}
\label{sec:conclusion}

We have presented a simple extension to semi-global matching (SGM)
that allows surface orientation priors to be incorporated as soft
constraints. Using priors derived from stereo matching at coarser
resolution, our SGM-P method consistently yields improved accuracy for
challenging indoor scenes that contain slanted weakly-textured
surfaces.  We also demonstrate the potential of orientation priors
derived from single images.  Our analysis involving oracle priors
demonstrates the potential for large performance gains.

Avenues for future work includes recovering more accurate orientation
priors, possibly via semantic analysis~\cite{bansal-cvpr2016}
or
%, room layout estimation based on Manhattan-world
%priors~\cite{ramalingam2013}, and
revisiting binocular photometric
stereo~\cite{du2011}. Combining orientation priors with depth priors,
obtained either from coarse resolution or via commodity depth sensors, is
also worth exploring. 
Finally, it might be possible to extend our method to
other MRF optimization frameworks with first-order smoothness terms.

{\small
\bibliographystyle{ieee}
\bibliography{bib}
}

\newpage
\clearpage

% abbreviation for bold face (overides underline...)
%\renewcommand{\b}[1]{\textbf{#1}}

%\threedvfinalcopy % *** Uncomment this line for the final submission

%\def\threedvPaperID{56} % *** Enter the 3DV Paper ID here
%\def\httilde{\mbox{\tt\raisebox{-.5ex}{\symbol{126}}}}

\setcounter{page}{1}
\setcounter{section}{0}
\setcounter{figure}{0}
\setcounter{table}{0}
\setcounter{equation}{0}
\setcounter{footnote}{0}
\renewcommand{\thepage}{A\arabic{page}}
\renewcommand{\thesection}{\Alph{section}}
\renewcommand{\thefigure}{A\arabic{figure}}
\renewcommand{\thetable}{A\arabic{table}}
\renewcommand{\theequation}{A\arabic{equation}}
\setstretch{0.99}

\def\httilde{\mbox{\tt\raisebox{-.5ex}{\symbol{126}}}}

\makeatletter
\def\@thanks{}
\makeatother

%\title{{Supplementary  Material for \\Semi-Global Stereo Matching with Surface Orientation Priors}}

\title{\mbox{}\\[-21mm]
SUPPLEMENTARY MATERIALS \\[2mm]
Semi-Global Stereo Matching with Surface Orientation Priors\\[-11mm] \mbox{}}

\author{Daniel Scharstein\\
Middlebury College\\
\and
Tatsunori Taniai\\
RIKEN AIP\\
\and
Sudipta N. Sinha\\
Microsoft Research
}

\maketitle

\section{Other matching costs and resolutions}

\begin{figure*}[b]
\newcommand{\ww}{3.35in}
\centering
\includegraphics[width=\ww]{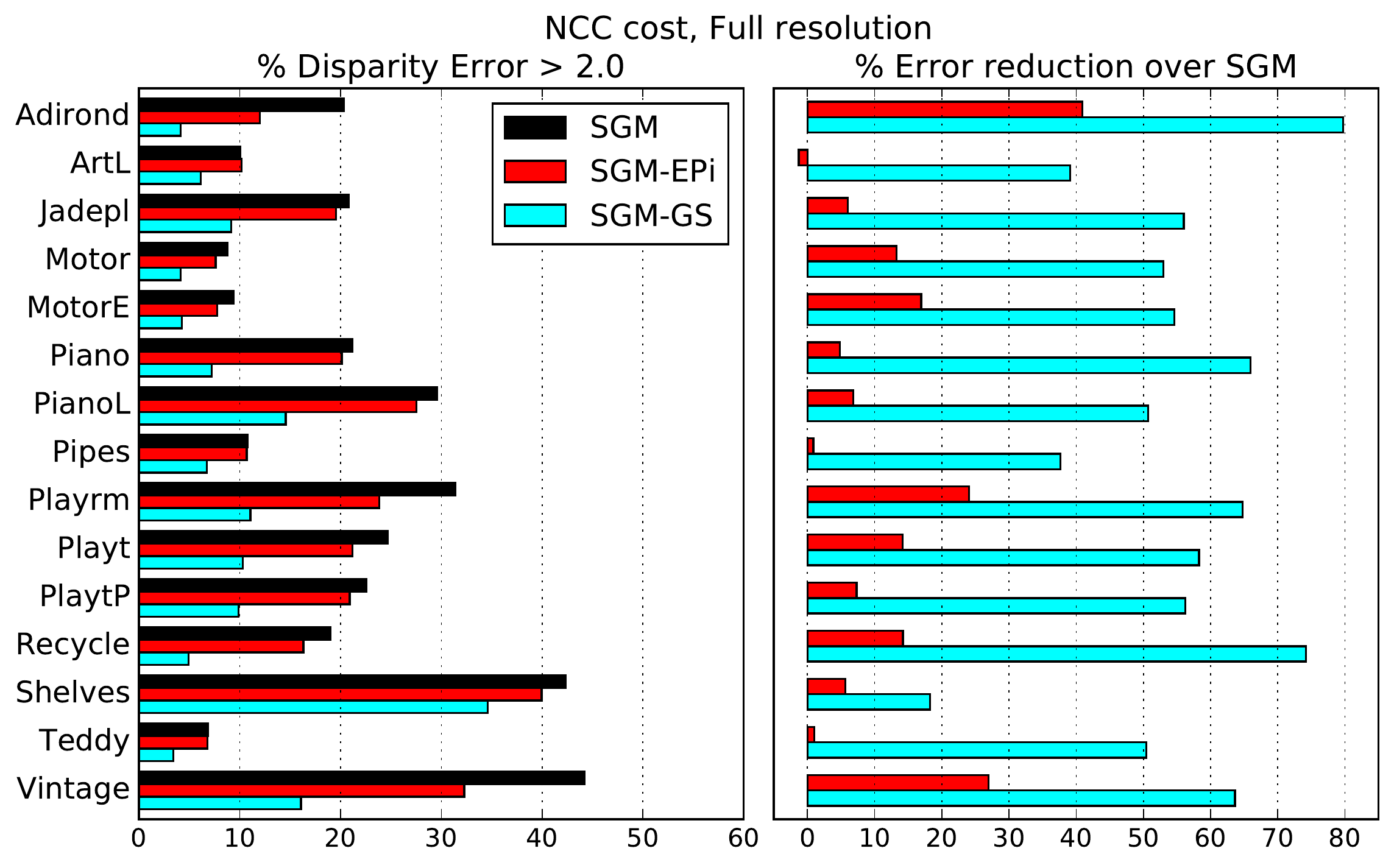} ~~~
\includegraphics[width=\ww]{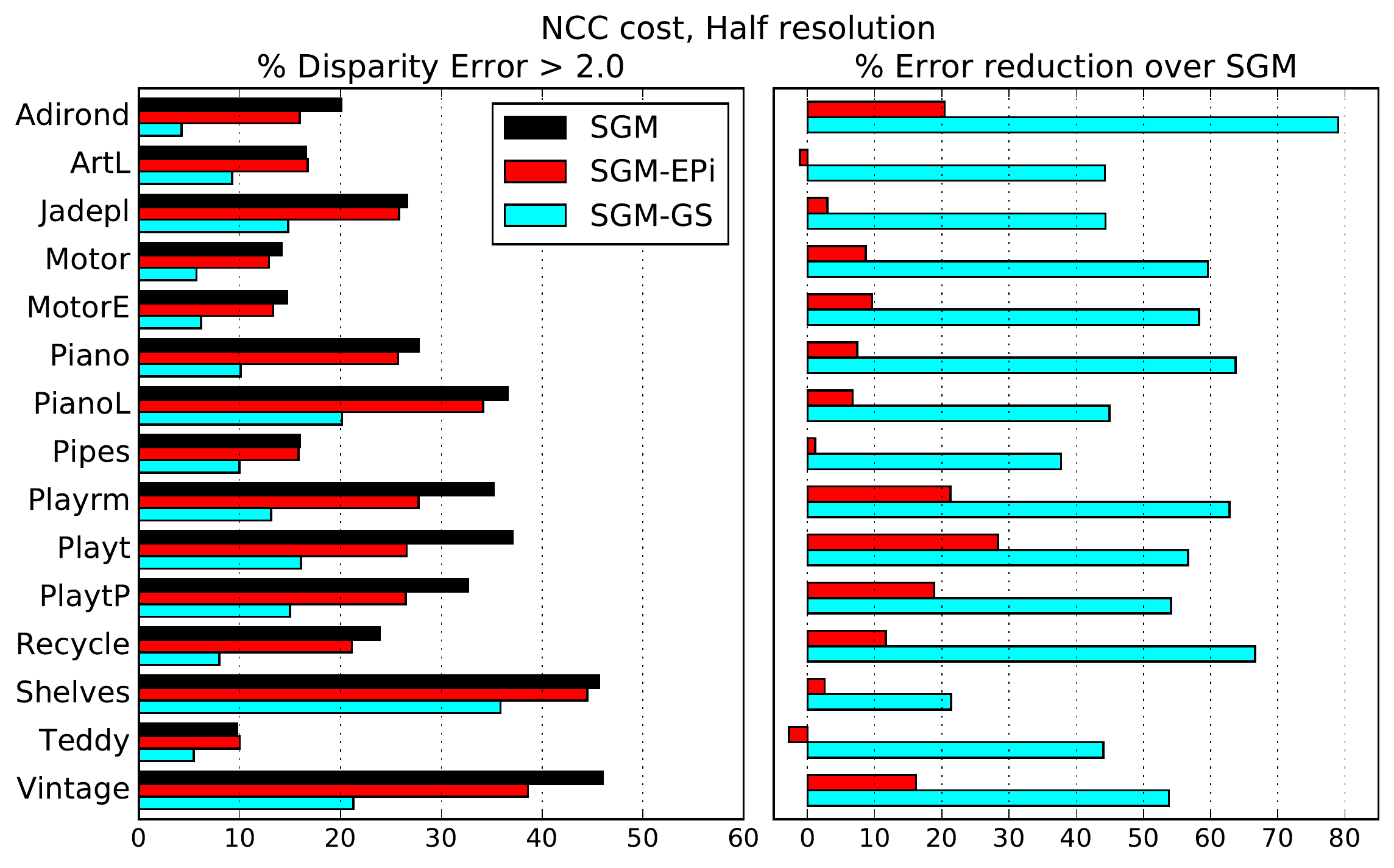} \\
\includegraphics[width=\ww]{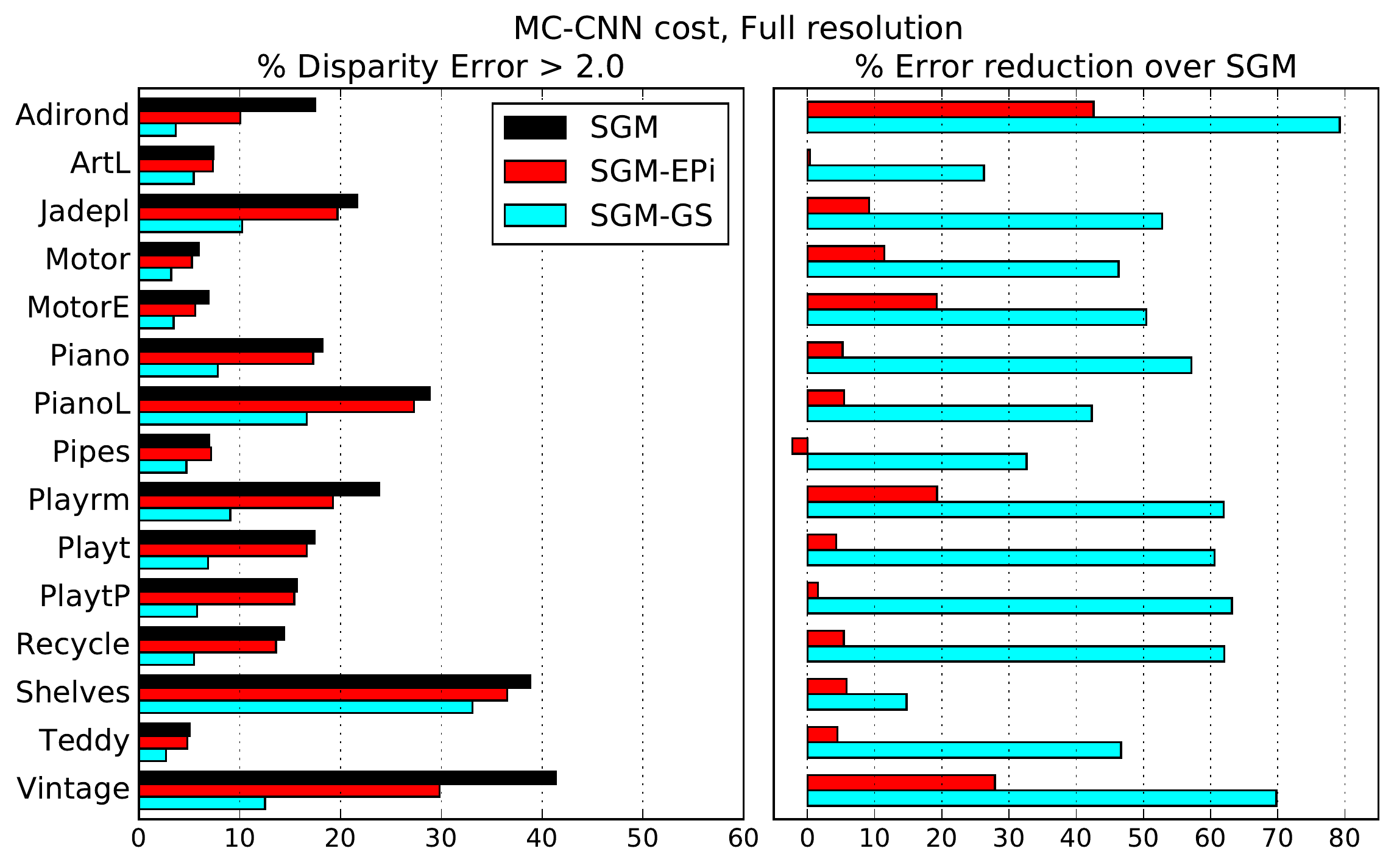} ~~~
\includegraphics[width=\ww]{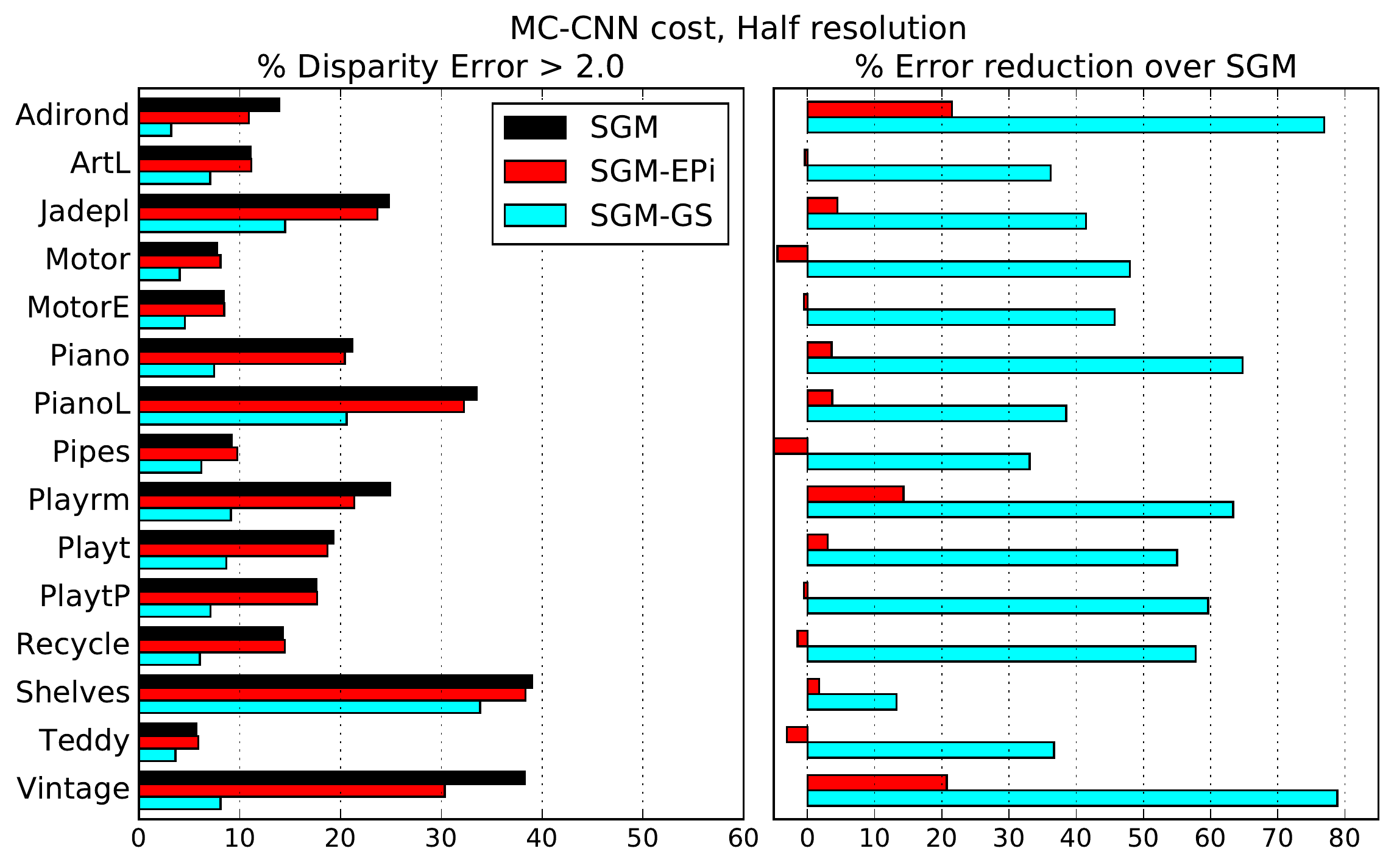}
\caption{Comparison of matching costs NCC (top) and MC-CNN (bottom),
at full (left) and half (right) resolution.% on the 15 Middlebury training pairs.
}
\label{fig:comparison}
\end{figure*}

In addition to NCC we also evaluate MC-CNN
%\cite{Zbontar2016}
[49]
as a matching
cost, on both full-resolution (6MP) and half-resolution (1.5MP)
versions of the Middlebury training images.  Fig.~\ref{fig:comparison}
shows barplots for baseline SGM, estimated priors SGM-EPi, and
ground-truth priors SGM-GS, on all four combinations of matching cost
and image resolution.  Table~\ref{tab:comparison} summarizes the
average performance gains.

For SGM-EPi, which estimates plane hypotheses from coarser matching
results, we run SGM with NCC matching costs at quarter resolution for
all of these combinations.  We found that MC-CNN is tuned for half
resolution and does not work well at quarter resolution.
\begin{table}
%{\small
\begin{tabular*}{3in}{ll|cc}
& & \multicolumn{2}{l}{\% error reduction} \\
Cost   & Resolution & SGM-EPi & SGM-GS \\ \hline
NCC    & Full  &   12.1  & 54.8 \\
       & Half  &   10.1  & 52.8 \\ \hline
MC-CNN & Full  &   10.7  & 51.1 \\
       & Half  &  ~~3.8  & 50.0 \\
\end{tabular*}
\vspace{1mm}
%}
\caption{Average performance gains for different matching costs and
image resolutions.}
\label{tab:comparison}
\end{table}
Fig.~\ref{fig:comparison} and Table~\ref{tab:comparison} show that
SGM-EPi yields comparable average performance gains of 10-12\% on all
combinations except for MC-CNN at half resolution, which overall
yields the lowest errors and thus smaller average gains.  However,
individual gains on difficult scenes such as Adirondack and Vintage
remain high across all combinations.  The oracle prior SGM-GS performs
well across all combinations, with average error reduction of at least
50\%.  Overall, these results demonstrate that our SGM-P method has
great potential independent of matching cost.

\begin{figure*}
\centering
\includegraphics[width=6.9in]{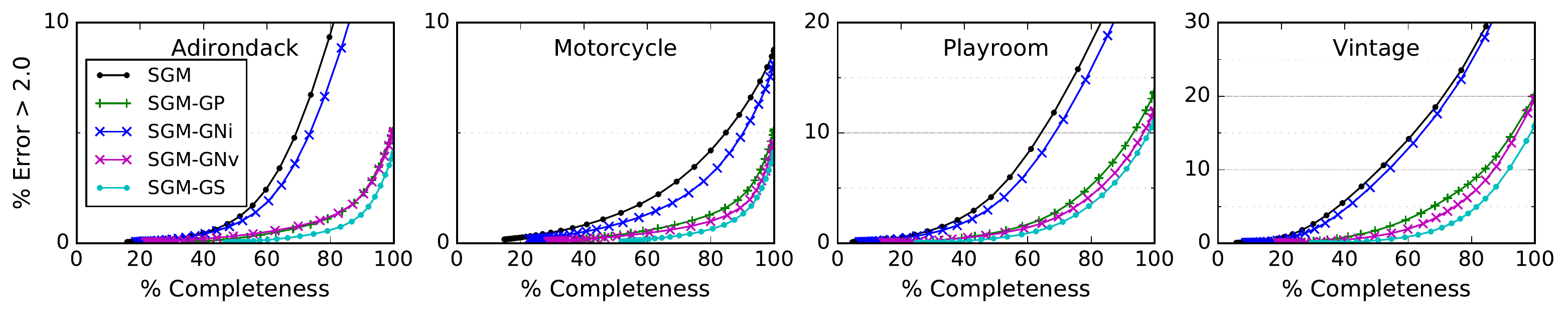}
\caption{Comparison of oracle orientation priors
SGM-GS (true surface), SGM-GP (planar approximation),
SGM-GNv (3D prior derived from true surface normals), and
SGM-GNi (2D ``strawman'' prior derived from true surface normals).
}
\label{fig:plots3}
\end{figure*}

\begin{figure*}[t]
\newcommand{\ww}{1.31in}
\centering
\begin{tabular}{@{}c@{~~}c@{~~}c@{~~}c@{~~}c@{}}
\includegraphics[width=\ww]{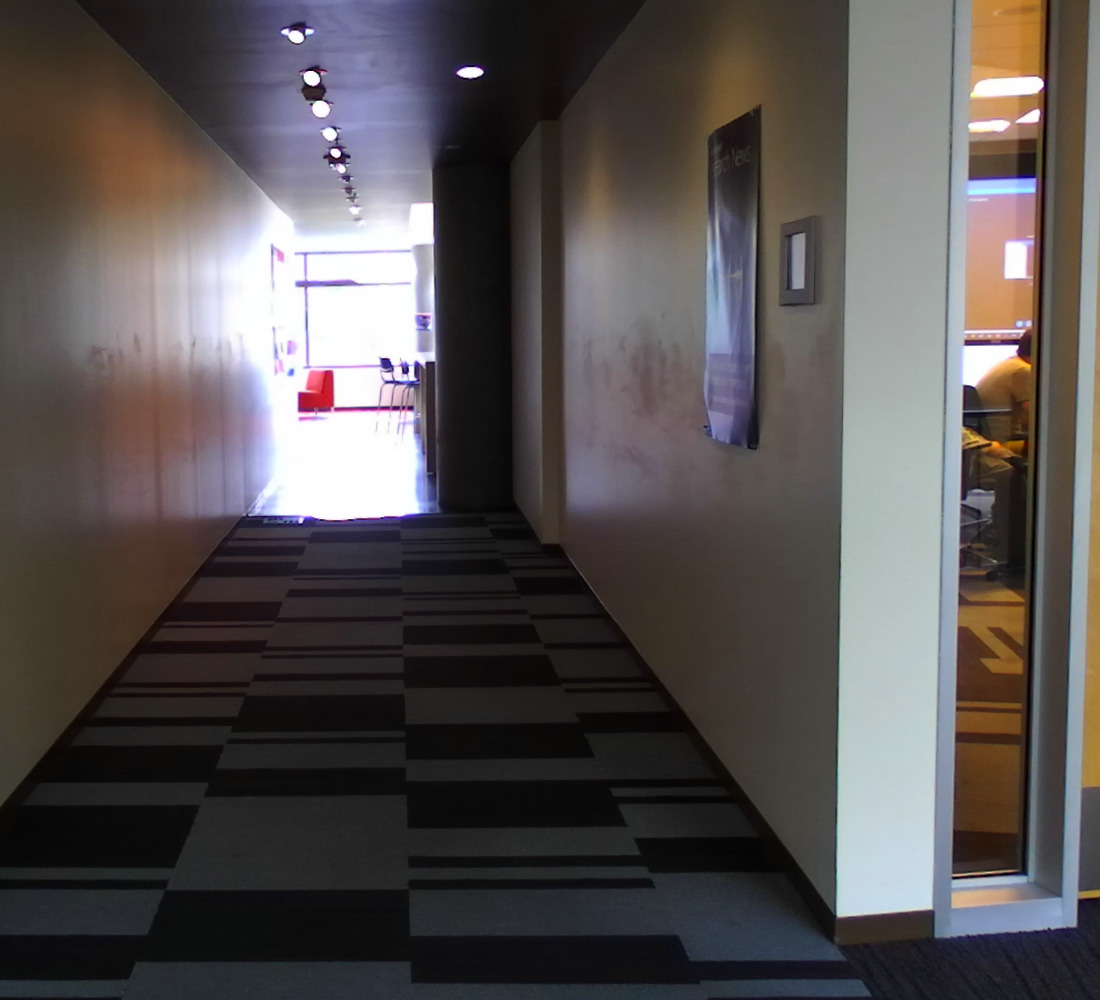} &
\includegraphics[width=\ww]{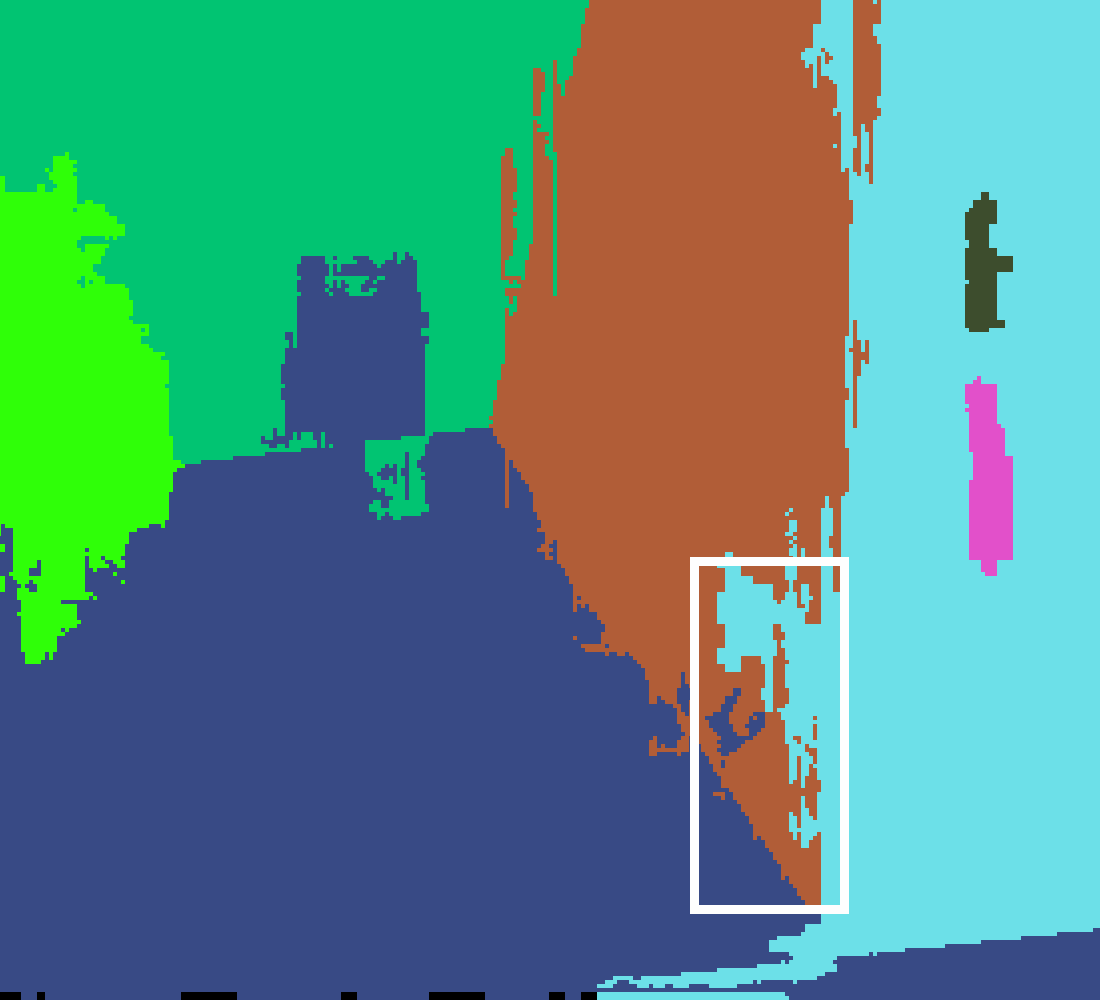} &
\includegraphics[width=\ww]{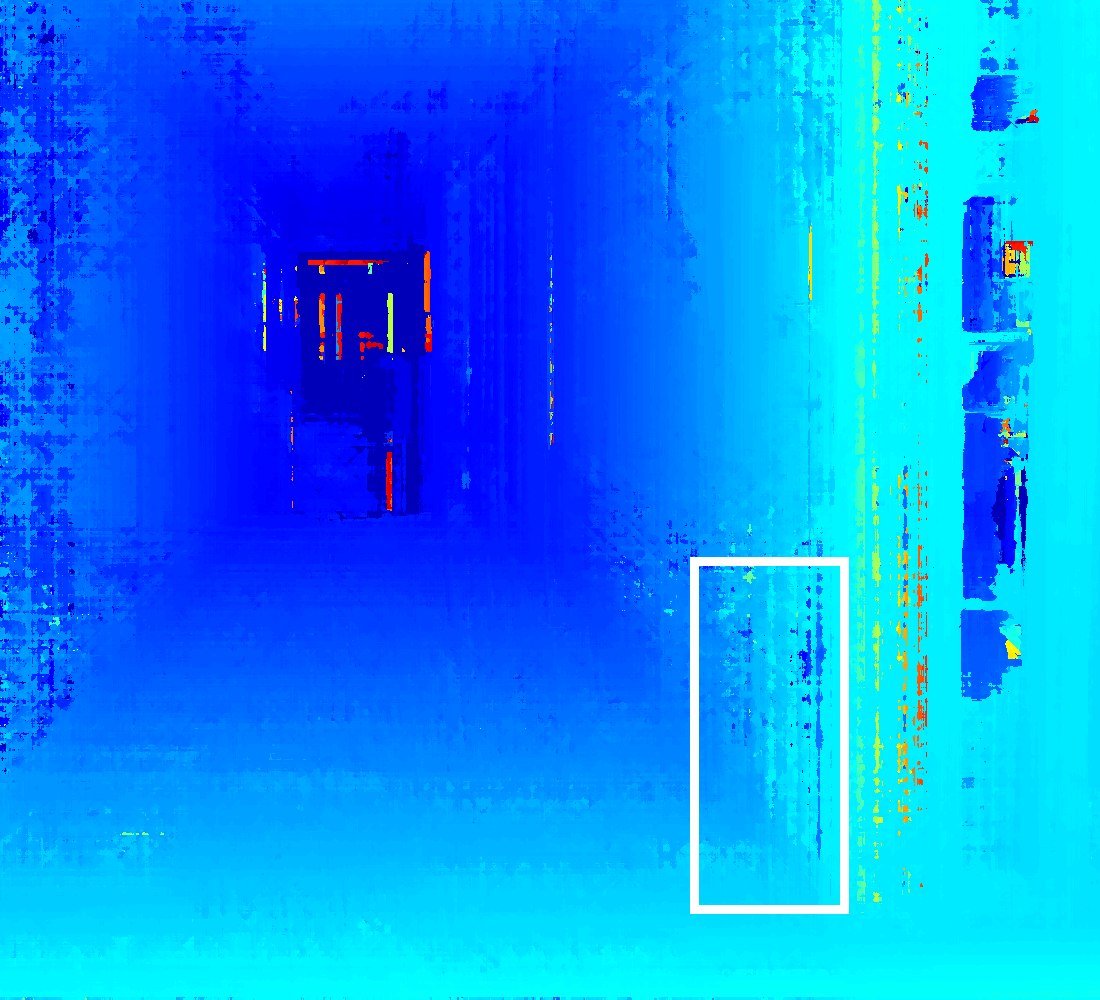} &
\includegraphics[width=\ww]{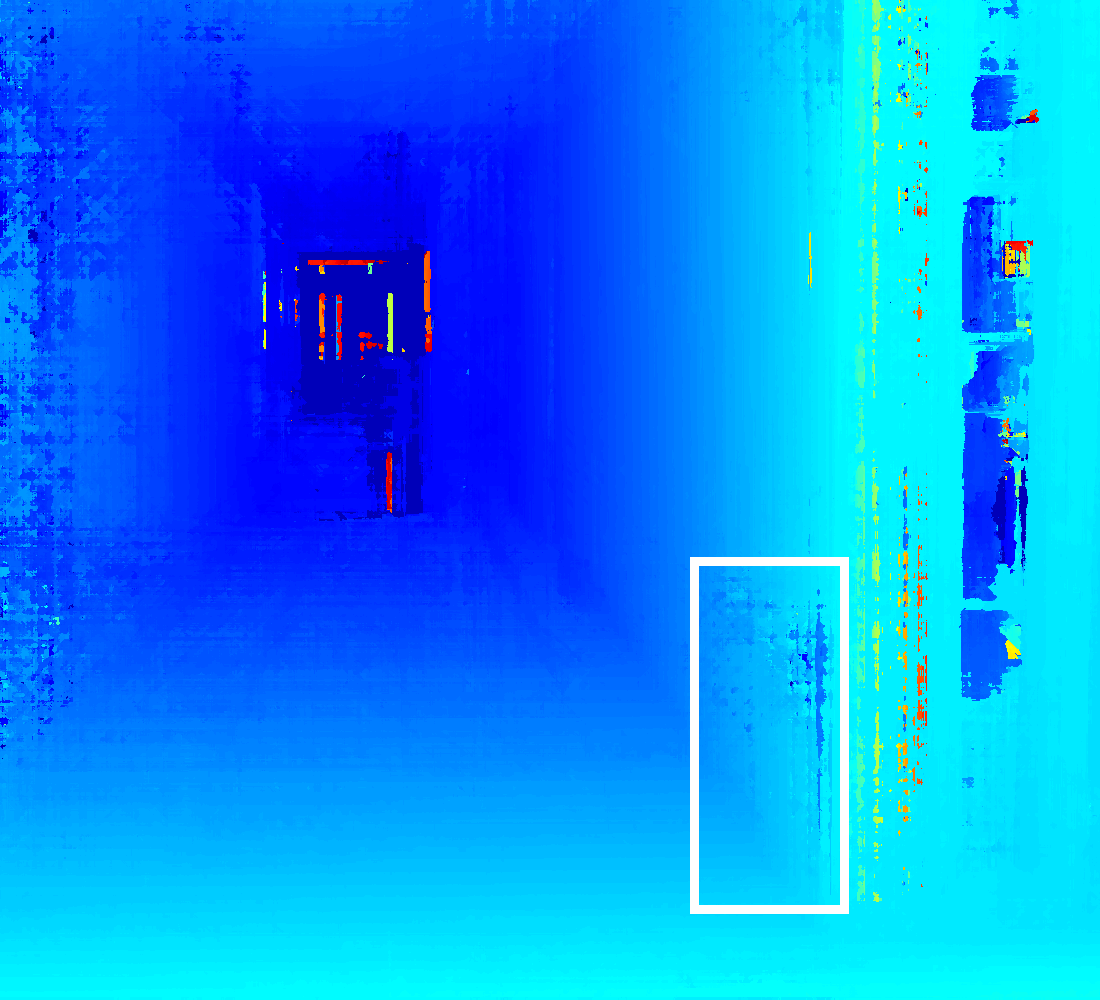} &
\includegraphics[width=\ww]{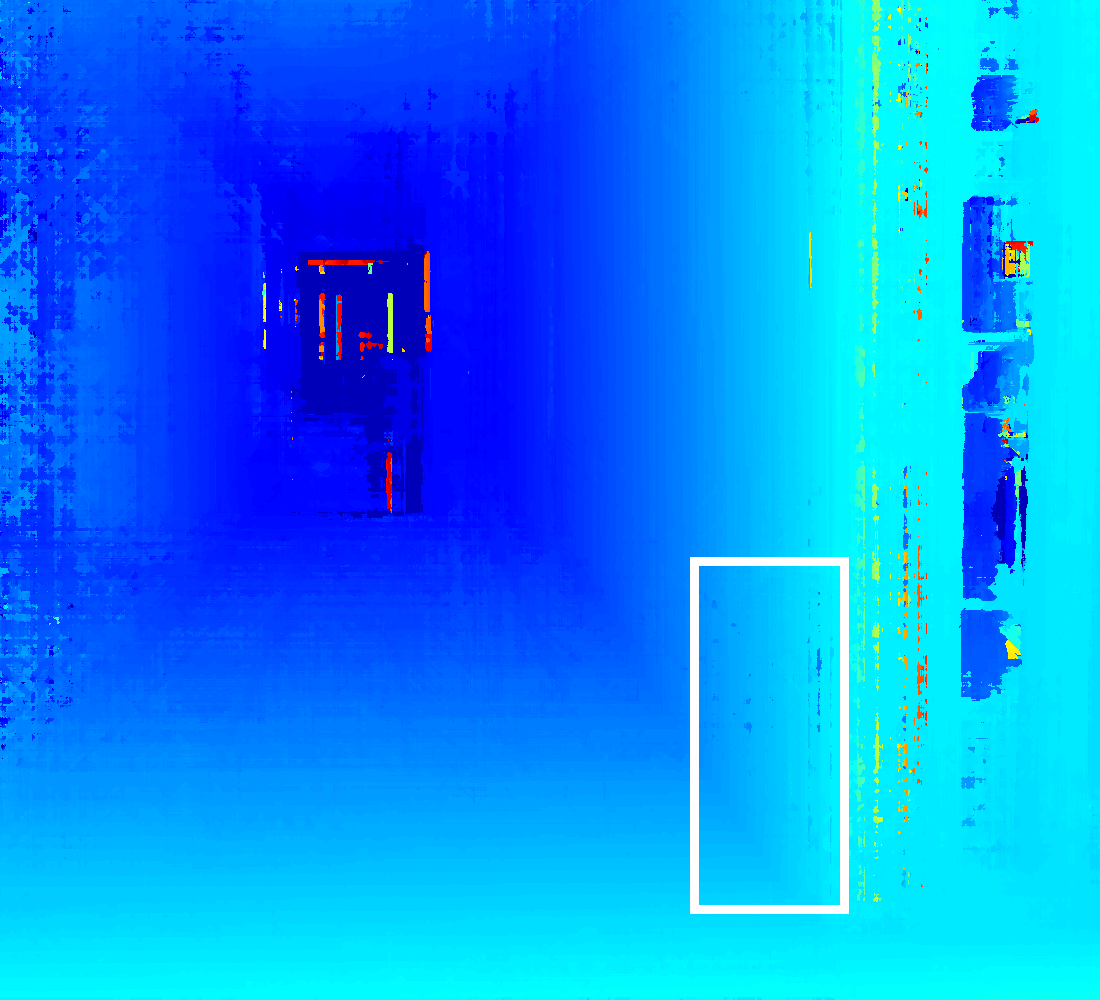} \\[-1mm]
\footnotesize (a) Input image &
\footnotesize (b) Plane labels &
\footnotesize (c) SGM &
\footnotesize (d) SGM-EPi &
\footnotesize (e) SGM-EPv \\[1mm]
\end{tabular}
\caption{Qualitative results using estimated priors.
(a) Challenging image pair.% taken with an off-the-shelf stereo camera.
(b) Planar priors extracted from noisy matching results at lower
resolution.  (c-e) Comparison of baseline (SGM) with estimated 2D
priors (SGM-EPi) and 3D priors (SGM-EPv).  Despite the low quality of
the estimated planes, the priors result in significantly cleaner
disparity maps compared to the baseline.  In addition, SGM-EPv yields
cleaner surface transitions (``creases'') in the presence of noisy labels
and/or competing priors, for instance near the front edge of the right corridor wall
(highlighted).}
\label{fig:2dvs3dimages}
\end{figure*}

\section{Oracle surface normal priors}

To allow an accurate comparison of the achievable benefit of priors
derived from surface normals (as opposed to the surfaces directly) we
include additional experiments using oracle priors derived from
ground-truth data.

In order to compute oracle priors from ground-truth surface normals
(SGM-GNv),  we integrate these normals in
scene space under a perspective projection model using a least-squares
approach
%\cite{tankus2005}.
[45].
The result is a $z$-surface in world
coordinates, initially at an arbitrary depth.
We implement the 2D integration step using a sparse solver employing
conjugate gradients. We divide the image into a coarse grid and
independently integrate a surface in each grid cell, arbitrarily
fixing one depth value in each cell.
In order for the integration to succeed, it is crucial that the
discontinuities in the normal map are known. Otherwise any integration
method, including our least-squares approach, will not produce a
locally accurate $z$-surface.

Recall from Section~3.5 in the paper that a planar surface patch with
known normal but unknown depth yields a family of planar
disparity surfaces whose orientation depends on $z$, resulting in a 3D
prior SGM-GNv. To investigate the importance of modeling this depth
dependance, we compare the accurate 3D version SGM-GNv with an
(inaccurate) 2D version SGM-GNi which we obtain by integrating the
surface normals in scene space using an arbitrary starting depth, and
using the resulting disparity surface as a 2D prior.  We compare both
the accurate 3D version SGM-GNv and the 2D ``strawman'' SGM-GNi with the
ground-truth surface prior SGM-GS and its piecewise-planar
approximation SGM-GP.  Fig.~\ref{fig:plots3} shows the
performance of these variants.
%, as well as the baseline SGM method,
%on the four image pairs also used in the paper.
It can be observed that the accurate normal prior SMG-GNv is close to
the upper bound SGM-GS, often outperforming the planar approximation
SGM-GP, while the strawman SGM-GNi performs much worse.  The
difference between SGM-GNv and SGM-GNi is less pronounced on other
image pairs where fewer slanted surfaces are present, or where the
single integration result coincides with a large actual surface in the
scene.
%, such as ground plane.

\section{Estimated 2D vs.~3D priors}

Recall from Table 2 in the paper that estimated 3D priors (SGM-EPv)
perform quantitatively slightly better than 2D priors (SGM-EPi).
Fig.~\ref{fig:2dvs3dimages} shows a qualitative example illustrating
why 3D priors are advantageous.  It can be seen that both SGM-EP
methods clearly produce much cleaner surfaces than the baseline SGM
algorithm.  In addition, SGM-EPv produces smoother results especially
near plane transitions at discontinuities and orientation changes
(``creases'') between planes.  Since SGM-EPi only has a single
orientation prior per pixel, its performance degrades when the
pixel-to-plane labeling is noisy or incomplete. SGM-EPv allows
multiple overlapping disparity hypotheses and is thus more robust in
the presence of noisy labels.

\end{document}